%% file: main.tex
\documentclass{article} 
\usepackage{iclr2027_conference,times}

\input{math_commands.tex}

\usepackage{amsmath}
\usepackage{amssymb}
\DeclareMathSizes{5.5}{5}{5}{5}
\DeclareMathSizes{7.5}{7}{5}{5}
\usepackage{microtype}
\usepackage{graphicx}
\usepackage{wrapfig}
\usepackage{hyperref}
\usepackage{url}
\title{\ours: Binocular Vision-Language \\ Models for Metric 3D Spatial Reasoning}

\author{Chi Yan$^{1,2,*}$\hspace{0.55em}Guoyang Zhao$^{2,*}$\hspace{0.55em}Wentao Yuan$^{2}$\hspace{0.55em}%
Yixiao Ge$^{2}$\hspace{0.55em}Yanning Zhou$^{2,\dagger}$\hspace{0.55em}Dan Xu$^{1,\dagger}$\\
\normalfont $^{1}$\,The Hong Kong University of Science and Technology (HKUST)\\
\normalfont $^{2}$\,XPENG Robotics\\
\normalfont\texttt{\{cyanao,danxu\}@cse.ust.hk}\\
\normalfont\small $^{*}$\,Equal contribution.\quad $^{\dagger}$\,Corresponding authors.
}

\iclrfinalcopy 

\begin{document}

\maketitle
\fancyhead{}
\renewcommand{\headrulewidth}{0pt}

\vspace{-2\baselineskip}
\input{sec/teaser}
\input{sec/0_abs}
\input{sec/1_intro}
\input{sec/2_related}

\input{sec/3_data_benchmark}
\input{sec/4_method}
\input{sec/5_exp}
\input{sec/6_conclusion}

\bibliographystyle{iclr2027_conference}
\bibliography{iclr2027_conference}

\raggedbottom
\input{appendix/detail}

\end{document}

%% file: math_commands.tex
\usepackage{amsmath,amsfonts,bm}

\usepackage{hyperref}
\usepackage{cleveref}
\usepackage{booktabs}
\usepackage{multirow}
\usepackage{xcolor}
\usepackage{colortbl}
\usepackage{xspace}
\newcommand{\ours}{StereoVLM\xspace}

\definecolor{stereogeometry}{RGB}{45, 85, 125}
\definecolor{stereosemantics}{RGB}{40, 110, 90}
\definecolor{stereooursbg}{gray}{0.93}
\makeatletter
\newcommand{\stereogap}[1]{%
  \cellcolor{stereooursbg}%
  \g@addto@macro\CT@cell@color{\@tempdimb#1\relax}%
}
\makeatother

\def\eqref#1{equation~\ref{#1}}

\def\1{\bm{1}}

\DeclareMathAlphabet{\mathsfit}{\encodingdefault}{\sfdefault}{m}{sl}
\SetMathAlphabet{\mathsfit}{bold}{\encodingdefault}{\sfdefault}{bx}{n}



%% file: sec/teaser.tex
\begin{figure}[!ht]
    \centering
    \includegraphics[width=\linewidth]{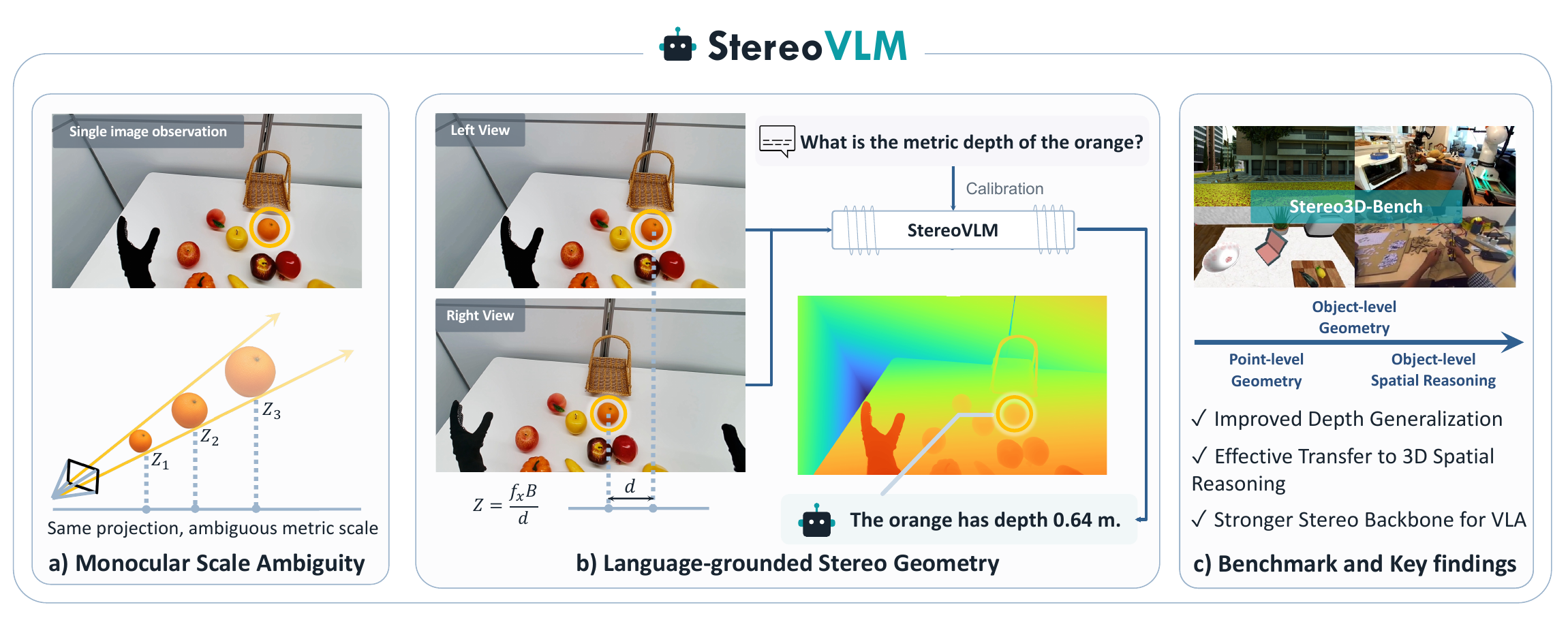}
    \vspace{-5ex}
    \caption{StereoVLM: from binocular geometry to spatial understanding and action.
    (a) Monocular observations leave metric scale ambiguous.
    (b) An illustrative example of language-grounded metric reasoning from
    calibrated stereo observations.
    (c) Stereo3D-Bench evaluates spatial understanding at three levels.
    Disparity supervision improves depth generalization, while hierarchical geometric
    supervision benefits broader 3D spatial reasoning and downstream VLA performance.}
    \label{fig:teaser}
\end{figure}

%% file: sec/0_abs.tex
\begin{abstract}
Vision-language models (VLMs) exhibit strong semantic understanding but remain limited in their ability to reason about metric 3D space in the physical world. Existing VLMs typically infer 3D scene structure from one or more images without known camera calibration or cross-view geometric constraints.
This makes metric 3D inference inherently ambiguous and leaves it unclear whether their spatial predictions are grounded in visual evidence or driven primarily by semantic priors. In contrast, calibrated stereo provides explicit cross-view geometry that resolves this ambiguity and offers a more reliable foundation for metric 3D perception and geometric generalization.
We investigate how VLMs can learn transferable 3D representations directly from calibrated binocular observations.
To this end, we introduce \textbf{\ours}, the first geometry-aware binocular VLM for metric 3D spatial reasoning.
\ours follows a hierarchical learning paradigm spanning from cross-view correspondence and disparity estimation to calibration-conditioned metric depth perception and semantic spatial reasoning. 
We further find that disparity supervision generalizes substantially better than direct binocular depth supervision on unseen environments.
To evaluate these capabilities, we introduce \textbf{Stereo3D-Bench}, a comprehensive benchmark across three levels, from point-level geometry to object-level spatial reasoning. 
\ours-8B achieves 95.17\% depth accuracy ($\delta_1$) and a 66.98 average object-level score on Stereo3D-Bench.
Intervention results further indicate that the model's performance on Stereo3D-Bench benefits from valid cross-view geometry and cannot be explained solely by additional visual tokens or semantic priors.
Hierarchical geometric supervision also yields average gains of 3.54 points across five spatial benchmarks and 4.3 percentage points in RoboCasa manipulation success.
\end{abstract}

%% file: sec/1_intro.tex
\section{Introduction}
\label{sec:intro}
\vspace{-3pt}

Vision-Language Models (VLMs) have demonstrated remarkable capabilities in understanding and reasoning about visual content in the digital world~\citep{alayrac2022flamingo}. Building upon these advances, recent efforts have begun to extend such capabilities beyond the digital realm into the physical world, positioning VLMs as a key foundation for embodied AI~\citep{driess2023palm, kim2024openvla, brohan2023rt, zhou2026si0, zhao2026geoprop}.
However, extending VLMs from the digital to the physical world requires more than semantic understanding: it demands geometrically grounded spatial intelligence for reasoning and acting in 3D environments.

To extend VLMs beyond semantic understanding, recent studies have explored various ways to enhance their spatial reasoning capabilities.
SpatialVLM~\citep{chen2024spatialvlm} learns qualitative and quantitative spatial concepts from large-scale spatial question-answering data, while SpatialRGPT~\citep{cheng2024spatialrgpt} augments region-level representations with monocular depth cues for object-centric reasoning. 3D-LLM~\citep{hong20233dllm}, in contrast, grounds language models in reconstructed 3D scene representations. Despite these advances, spatial benchmarks such as VSI-Bench reveal persistent limitations in current models' ability to construct and reason about 3D space from visual observations~\citep{yang2025thinking}. 
Moreover, the geometric information available to these models is typically inferred from monocular observations or supplied through estimated depth and reconstructed 3D representations, making it difficult to determine whether their spatial predictions are grounded in geometric evidence directly available from visual observations or primarily reflect learned priors—a concern reinforced by evidence-absent evaluations showing that VLMs can exploit scene- and category-level priors~\citep{zhang2026revsi}.
Calibrated stereo provides explicit geometric constraints for metric 3D perception: cross-view correspondences determine disparity, which can be converted into metric depth using known camera intrinsics and baseline. Yet, providing binocular observations alone does not guarantee that a VLM will acquire geometric representations that generalize beyond the training distribution. Existing stereo methods primarily focus on geometric estimation, while VLMs typically treat paired images as generic visual inputs without explicitly learning the geometric relations between them. \textbf{This raises a central question:} \emph{How can a VLM learn transferable representations from binocular geometry that support metric 3D perception and high-level spatial reasoning?} 
To investigate this question, we introduce StereoVLM, a geometry-aware binocular VLM designed to learn transferable 3D representations directly from calibrated stereo observations.

To connect binocular geometry with spatial understanding, we organize the training supervision into a hierarchy of tasks spanning pixel-level correspondence, disparity estimation, calibration-conditioned metric depth, and object-level spatial reasoning. This hierarchy progressively connects cross-view matching to image-space geometric measurement and then to metric-scale estimation, before extending the learned geometry to language-referred objects and task-relevant 3D grounding.
Importantly, we find that disparity provides a substantially more generalizable supervision signal than direct binocular depth on unseen environments, motivating its role as an intermediate geometric representation between cross-view correspondence and metric depth. We further establish Stereo3D-Bench to evaluate this progression across point-level geometry, object-level geometry, and object-level spatial reasoning. Beyond conventional task accuracy, controlled right-view interventions and geometric out-of-distribution evaluation test whether model predictions genuinely depend on valid cross-view geometry. Compared with matched training without stereo instructions, hierarchical geometric supervision raises the five-benchmark spatial-reasoning average by 3.54 points and RoboCasa~\citep{nasiriany2024robocasa} manipulation success by 4.3 percentage points.

Our contributions are:
\vspace{-4pt}
\begin{itemize}
    \item We introduce StereoVLM, the first geometry-aware binocular VLM for metric 3D spatial reasoning, with hierarchical supervision that connects cross-view correspondence and disparity to calibration-conditioned metric depth and language-grounded spatial reasoning, yielding transferable 3D spatial representations.
    \vspace{-4pt}
        \item We establish Stereo3D-Bench, a comprehensive benchmark spanning point-level geometry, object-level geometry, and object-level spatial reasoning, together with a verifiable evaluation protocol that uses controlled input interventions and geometric out-of-distribution evaluation to test whether predictions genuinely rely on binocular geometry.
    \vspace{-12pt}
    \item We demonstrate that disparity provides a more generalizable geometric supervision signal than direct binocular depth on unseen environments, while hierarchical geometric supervision benefits broader spatial reasoning and downstream embodied control.
\end{itemize}
\vspace{-8pt}

%% file: sec/2_related.tex
\vspace{-2pt}
\section{Related Work}
\label{sec:related}

\vspace{-6pt}
\paragraph{Spatial reasoning in vision-language models}
Existing studies have enhanced spatial reasoning in VLMs through spatial instruction tuning~\citep{chen2024spatialvlm,cao2026spatialspeak}, depth-aware visual representations~\citep{cheng2024spatialrgpt}, and 3D grounding~\citep{hong20233dllm,wang2025n3d}. Meanwhile, benchmarks such as VSI-Bench~\citep{yang2025thinking} reveal that current multimodal models still struggle with metric 3D understanding and spatial reasoning. Recent works explore metric depth prediction in VLMs, including queried-point prediction in DepthLM~\citep{yeh2026depthlm} and dense prediction in DepthVLM~\citep{yu2026unlocking}. VLM$^3$~\citep{cai2026vlm3} covers metric depth, cross-image correspondence, camera pose estimation, and object-level 3D understanding through separate task-specific training. Joint learning across low-level geometry and high-level semantic spatial understanding remains less explored.

\vspace{-1ex}
\paragraph{Stereo vision and embodied intelligence}
Stereo vision recovers scene geometry by establishing cross-view correspondence and converting disparity into depth using camera calibration. Modern stereo methods, including RAFT-Stereo~\citep{lipson2021raft}, IGEV-Stereo~\citep{xu2023iterative},  FoundationStereo~\citep{wen2025foundationstereo} and S2M2~\citep{min2025s}, employ correlation or geometry volumes to achieve accurate and generalizable disparity estimation. Recent work has begun incorporating binocular sensing into embodied systems. StereoVLA~\citep{deng2025stereovla} fuses stereo-geometric and monocular-semantic features and adds auxiliary depth and camera-parameter objectives for VLA manipulation, while StereoPolicy~\citep{han2026stereopolicy} uses cross-view attention to adapt diffusion policies and pretrained VLA policies without explicit 3D reconstruction. EATR-Stereo~\citep{wu2026eatr} further routes paired-view evidence according to proprioceptive state for long-horizon humanoid control, and VLBiMan~\citep{zhou2026vlbiman} combines vision-language grounding with calibrated stereo reconstruction in a modular one-shot bimanual system. These methods demonstrate the value of binocular cues, with a primary focus on geometric estimation or embodied action. Our work bridges embodied stereo sensing and VLM spatial reasoning through a stereo VLM jointly trained on cross-view correspondence, disparity estimation, calibration-conditioned metric depth perception, and high-level semantic spatial reasoning.

%% file: sec/3_data_benchmark.tex
\section{Stereo3D-Bench}
\label{sec:data}

\begin{figure}[t]
    \centering
    \vspace{-1ex}
    \includegraphics[width=\linewidth]
    {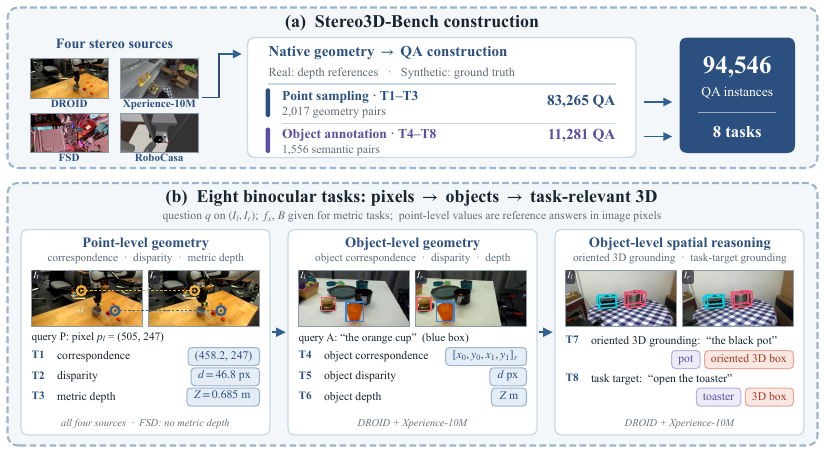}
    \vspace{-4.5ex}
    \caption{Construction and task hierarchy of \textbf{Stereo3D-Bench}.
    (a) The main split is constructed from four stereo data sources.
    Independently selected point- and object-level subsets contain 2,017
    and 1,556 stereo pairs, respectively, yielding 94,546 question--answer instances.
    (b) Representative queries and answer formats for eight tasks spanning
    point-level geometry, object-level geometry, and object-level spatial reasoning.}
    \vspace{-1ex}
    \label{fig:benchmark_overview}
\end{figure}

\vspace{-1ex}
Stereo3D-Bench evaluates whether binocular evidence supporting metric
geometry also transfers to language-grounded understanding of objects and
task-relevant 3D structure. Its main split draws on four data sources---DROID, Xperience-10M, FSD,
and RoboCasa---and covers all eight tasks. An out-of-domain (OOD) split
uses independently collected in-house data to evaluate point-level
correspondence, disparity, and depth. \Cref{fig:benchmark_overview}
illustrates the benchmark construction and task hierarchy. Controlled right-view
interventions further test whether predictions depend on valid cross-view
correspondence rather than merely on the presence of a second image. The
supplementary material details data and question construction
(\cref{sec:sup_data}), split and leakage controls
(\cref{sec:sup_splits}), and the evaluation protocol
(\cref{sec:sup_protocol}).

\subsection{Task Hierarchy}
\label{sec:benchmark_tasks}

Stereo3D-Bench comprises eight tasks across three levels: three point-level geometry tasks, three object-level geometry tasks, and two object-level spatial reasoning tasks. Each query is posed as a short language question on an ordered stereo pair.
Depending on the task, the model
returns a coordinate, a scalar, a 2D box, or an oriented 3D box with an object label.

\textbf{Point-level geometry.}
Given a queried pixel in the left image, the model predicts its corresponding location in the right image, its horizontal disparity, or its metric depth along the left-camera optical axis.
These tasks assess the model's ability to establish cross-view correspondence, estimate binocular disparity, and recover metric scene geometry from stereo observations.

\textbf{Object-level geometry.}
The same three geometric tasks are applied to an object specified by
language. Object correspondence returns its matching bounding box in the
right image, object disparity predicts its representative disparity, and
object depth predicts its representative metric depth along the left-camera optical axis.

\textbf{Object-level spatial reasoning.}
Oriented 3D grounding identifies a referred object and predicts its label
and oriented 3D bounding box in the rectified-left camera frame.
Task-target grounding instead receives a manipulation instruction,
identifies the object to be manipulated, and predicts that object's oriented 3D bounding box.

\vspace{-1ex}
Together, these tasks test whether a model can progress from matching
individual pixels, to associating geometry with named objects, and
to grounding task-relevant objects in metric 3D space.

\subsection{Data Sources and VQA Construction}
\label{sec:benchmark_data}

The main split combines real-world data from
DROID~\citep{khazatsky2024droid} and
Xperience-10M~\citep{xperience_10m} with synthetic data from
FSD~\citep{wen2025foundationstereo} and
RoboCasa~\citep{nasiriany2024robocasa}. For the synthetic sources, FSD
provides rendered ground-truth disparity, while RoboCasa provides
ground-truth disparity and depth directly from the simulator. FSD lacks
physical metric calibration and is therefore evaluated only for
correspondence and disparity.

For the real-world sources, DROID provides aligned sensor depth, while
Xperience-10M provides its officially released metric stereo-depth
annotations. The five object-level
tasks are evaluated on DROID and Xperience-10M.

\paragraph{Evaluation data.}
The main split uses held-out samples from the four sources above.
Training and evaluation data are separated before annotation generation.
DROID is split by scene (laboratory, robot, and collection date), and
Xperience-10M by recording UUID; all images and derived questions from
held-out groups are excluded from training. FSD evaluation frames are
excluded from training, while RoboCasa reserves five of its 24 task
identities for evaluation and excludes all associated samples from training.
To further assess geometric generalization to unseen scenes, we
independently collected an in-house OOD test set covering point-level
correspondence, disparity, and depth. Results are reported separately.
Split details are provided in \cref{sec:sup_splits}.

\paragraph{VQA construction.}
Using the source-specific geometric references described above, we
construct question-answer pairs for the tasks in
\cref{sec:benchmark_tasks}. Point-level queries are sampled from valid
image regions. For real-world sources, S2M2~\citep{min2025s} provides
an independent consistency check. A pixel $\mathbf{p}_l$ is eligible
only when both depth values are valid and satisfy
\begin{equation}
\frac{\left|Z_{\mathrm{S2M2}}(\mathbf{p}_l)
-Z_{\mathrm{native}}(\mathbf{p}_l)\right|}
{Z_{\mathrm{native}}(\mathbf{p}_l)} \leq 0.10,
\label{eq:benchmark_depth_agreement}
\end{equation}
where $Z_{\mathrm{S2M2}}$ is the depth recovered from S2M2 disparity
using camera calibration and $Z_{\mathrm{native}}$ is the source-native
depth reference. Further spatial filtering is detailed in
\cref{sec:sup_geometry}. S2M2 is used only for filtering; numerical
answers are derived from the source-native references.
For each stereo pair and applicable point-level task, we sample 120 points
and group them into 15 questions of eight points each. QA counts refer to
grouped questions rather than individual points.

For object-level queries, Qwen3~\citep{bai2025qwen3} proposes objects,
SAM3~\citep{carion2026sam} generates their masks, and
DAM~\citep{lian2025describe} provides region descriptions to reject
semantically inconsistent proposals. For each retained object $i$, we
back-project its valid pixels into a point cloud in the rectified-left
camera frame:
\begin{equation}
\mathcal{P}_i = \left\{
Z_{\mathrm{native}}(\mathbf{p})\,\mathbf{K}_l^{-1}\tilde{\mathbf{p}}
\;\middle|\; \mathbf{p}\in\mathcal{M}_i\cap\mathcal{V}
\right\},
\label{eq:benchmark_object_pointcloud}
\end{equation}
where $\mathcal{M}_i$ is the object's segmentation mask, $\mathcal{V}$
is the valid pixel set after all filtering steps, $\mathbf{K}_l$ is the
left-camera intrinsic matrix, and $\tilde{\mathbf{p}}=(u,v,1)^\top$
is the homogeneous pixel coordinate. We estimate object extent and
orientation in a gravity-aligned frame obtained from DROID camera
extrinsics or estimated for Xperience-10M, then transform the boxes back
to the rectified-left camera frame.
Task-specific templates convert the annotations into questions and
structured answers, with camera calibration included in metric-depth
queries. Both training and benchmark annotations undergo automatic
semantic filtering and box--mask reprojection checks; benchmark questions
additionally receive manual review.

Training follows the same question schemas but uses
FoundationStereo~\citep{wen2025foundationstereo} for numerical supervision.
Where source-native references are available, their agreement with
FoundationStereo identifies reliable training regions. FoundationStereo
is used neither to filter benchmark pixels nor to generate benchmark
answers.
Details are provided in \cref{sec:sup_data}.

\subsection{Binocular Controls}
\label{sec:benchmark_controls}

A correct prediction does not by itself demonstrate that the model uses
binocular evidence. We evaluate each query under five input
conditions while holding the checkpoint, question, calibration, and
decoding procedure fixed. \emph{Correct stereo}
$(\mathbf{I}_l,\mathbf{I}_r)$ defines the primary benchmark score; the
other four conditions are diagnostic controls.
\emph{Single left} supplies only $\mathbf{I}_l$, whereas
\emph{zero right} supplies
$(\mathbf{I}_l,\mathbf{0})$, where $\mathbf{0}$ is a zero-valued image
with the same dimensions as $\mathbf{I}_r$. These controls test whether
any improvement requires visual evidence from the right image rather than
merely a second image slot.
\emph{Duplicated left} supplies
$(\mathbf{I}_l,\mathbf{I}_l)$, removing the binocular displacement while
preserving the same scene content. \emph{Shuffled right} supplies
$(\mathbf{I}_l,\mathbf{I}'_r)$, where $\mathbf{I}'_r$ is taken from another
benchmark sample. It preserves a natural second image but breaks the
cross-view correspondence. Together, these controls test whether predictions
depend on the geometrically matching right view.

%% file: sec/4_method.tex
\section{StereoVLM}
\label{sec:method}

\begin{figure}[!t]
    \centering
    \includegraphics[width=\linewidth]{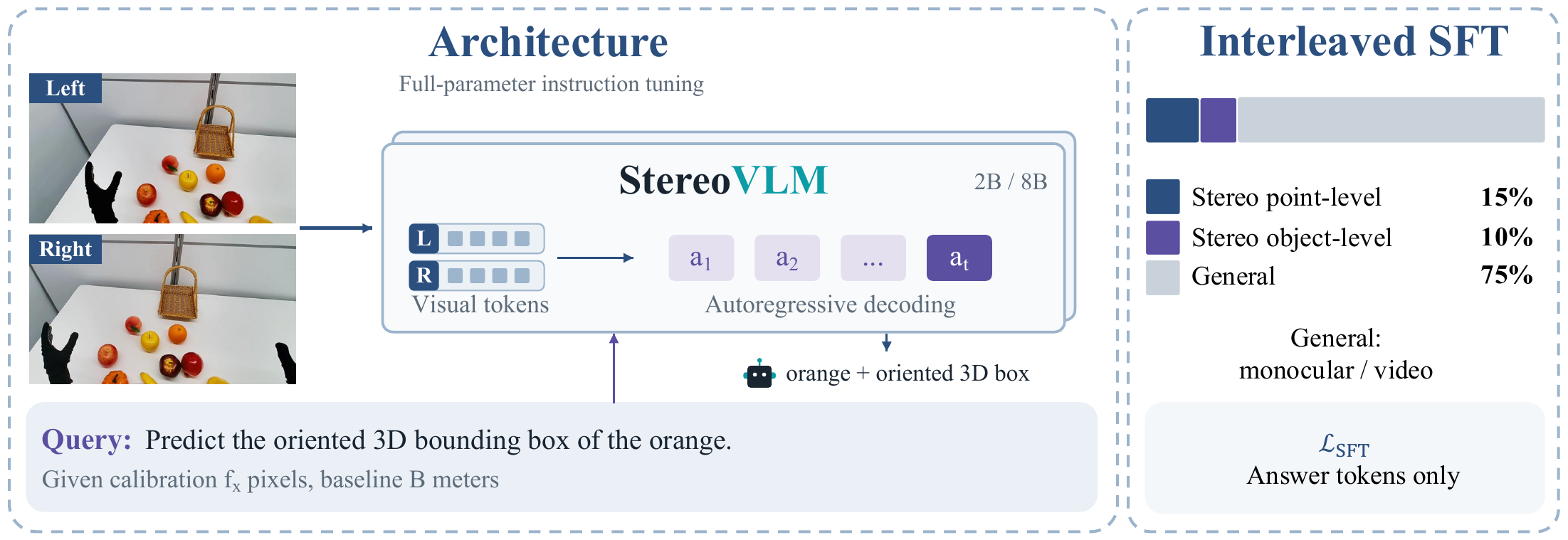}
    \vspace{-5ex}
    \caption{StereoVLM architecture and instruction tuning.
    StereoVLM jointly processes an ordered stereo pair and a textual query
    to generate answers autoregressively, with focal length $f_x$ and baseline
    $B$ provided for metric reasoning. Full-parameter instruction tuning
    interleaves point-level stereo instructions (15\%), object-level stereo
    instructions (10\%), and general monocular and video instructions (75\%).
    The SFT loss is applied only to answer tokens.}
    \label{fig:method_pipeline}
    \vspace{-2ex}
\end{figure}

\vspace{-1ex}
StereoVLM adapts a pretrained vision-language model to binocular spatial
reasoning through hierarchical instruction supervision. We first formulate
the model inputs and outputs, then describe how stereo geometry connects
the supervision targets, and finally introduce the joint training strategy.
\Cref{fig:method_pipeline} summarizes the architecture and training pipeline.

\subsection{Model Formulation}
\label{sec:stereo_formulation}

Given an ordered pair of rectified stereo images $(I_l,I_r)$ and a
question $q$, StereoVLM generates a textual answer $a$ describing the queried
geometric or spatial property:
\begin{equation}
\vspace{-2pt}
p_\theta(a\mid I_l,I_r,q)
=\prod_{t=1}^{T}p_\theta(a_t\mid I_l,I_r,q,a_{<t}),
\label{eq:stereo_prediction}
\vspace{-2pt}
\end{equation}
where $\theta$ denotes the model parameters, $T$ is the answer length,
and $a_t$ and $a_{<t}$ denote the current and preceding answer tokens,
respectively. The question specifies image points, a
referred object, or a manipulation instruction, and includes camera
calibration, including the horizontal focal length $f_x$ and stereo
baseline $B$, when metric geometry is required. These parameters allow
predictions to account for the physical scale of the stereo setup. Built
on Qwen3-VL-Instruct~\citep{bai2025qwen3}, StereoVLM processes both images
and the textual query through the pretrained vision-language architecture.
We adapt this model using hierarchical geometric supervision, as described
below.

\vspace{-1ex}
\subsection{Hierarchical Geometric Supervision}
\label{sec:geometry_supervision}

\paragraph{From correspondence to metric depth.}
We formulate point-level supervision as three complementary predictions
of the same underlying stereo geometry: correspondence, disparity, and
metric depth. Their targets are derived from a shared FoundationStereo
disparity estimate (\cref{sec:sup_geometry}), with camera calibration
providing the conversion to metric scale. For a queried pixel
$\mathbf{p}_l=(u_l,v_l)$ in $I_l$, its correspondence
$\mathbf{p}_r=(u_r,v_r)$ in $I_r$ satisfies $v_l\simeq v_r$, and
\begin{equation}
\vspace{-2pt}
d(\mathbf{p}_l)=u_l-u_r.
\label{eq:stereo_disparity}
\vspace{-2pt}
\end{equation}
When physical calibration is available, the depth target along the
left-camera optical axis is given by
\begin{equation}
\vspace{-2pt}
Z(\mathbf{p}_l)=\frac{f_xB}{d(\mathbf{p}_l)},
\qquad d(\mathbf{p}_l)>0.
\label{eq:disparity_to_depth}
\vspace{-2pt}
\end{equation}
Here, disparity $d$ and focal length $f_x$ are in pixels, while baseline $B$
and depth $Z$ are in meters.

We encode these targets as separate question-answer pairs and jointly
optimize the model across all three question types. Correspondence
supervision emphasizes cross-view localization, disparity supervision
quantifies image-space displacement, and depth supervision requires
interpreting that displacement under the supplied calibration. Deriving
all three targets from a common geometric reference preserves their
geometric relationship during target construction, while joint training
encourages the model to learn both cross-view matching and its connection
to metric scale.

\paragraph{Language-grounded object supervision.}
Point-level supervision captures local stereo geometry. To associate this
geometry with semantic entities and their spatial extent, we introduce
object-level supervision conditioned on linguistic references. Object
correspondence, disparity, and depth link the referred object to its
cross-view location and geometric measurements, while oriented 3D grounding
supervises its position, dimensions, and orientation in the rectified-left
camera frame. Task-target grounding further requires the model to identify
and localize the object specified by a manipulation instruction. Together,
these objectives extend supervision from local geometric estimation to
language-conditioned object localization and task-relevant 3D grounding.

\subsection{Joint Training Objective}
\label{sec:joint_training}

We jointly optimize StereoVLM over point-level geometry, object-level
grounding, and general multimodal instructions. All tasks share the
conditional answer-generation formulation in \cref{sec:stereo_formulation},
allowing geometric measurements and language-grounded spatial outputs to
be learned under a unified objective. Let $\mathcal{D}_{\mathrm{mix}}$
denote the resulting instruction distribution and $x$ the visual and
textual input, with $x=(I_l,I_r,q)$ for stereo examples. The training
objective is
\begin{equation}
\vspace{-2pt}
\mathcal{L}_{\mathrm{SFT}}(\theta)
=-\mathbb{E}_{(x,a)\sim\mathcal{D}_{\mathrm{mix}}}
\left[\sum_{t=1}^{T}\log p_\theta(a_t\mid x,a_{<t})\right].
\label{eq:instruction_loss}
\end{equation}
The loss is applied only to answer tokens. Point- and object-level
instructions are interleaved throughout training, so the supervision
hierarchy describes complementary levels of spatial understanding rather
than sequential optimization stages. Details are reported in \cref{sec:exp_setup}.
\vspace{-1.5ex}

%% file: sec/5_exp.tex
\section{Experiments}
\label{sec:exp}

\subsection{Experimental Setup}
\label{sec:exp_setup}

\paragraph{Model and training setup.}
We initialize \ours-2B and \ours-8B from the corresponding
Qwen3-VL-Instruct~\citep{bai2025qwen3} checkpoints. Unless otherwise
specified, our final models are instruction-tuned for 3,000 steps with
all parameters updated, using an initial learning rate of
$1\times10^{-5}$, a warmup ratio of $0.03$, and an effective global batch
size of 64. Stereo images are processed at a resolution of $800\times320$.
Stereo instructions account for 25\% of the sampled training mixture,
with 15\% assigned to point-level geometry and 10\% to object-level
supervision. The remaining 75\% comprises general visual-language,
grounding, spatial-reasoning, tracking, and robotics instructions.
Full data and optimization details are provided in \cref{sec:sup_training}.

\paragraph{Evaluation.}
We evaluate \ours{} on Stereo3D-Bench for binocular geometry and
object-level spatial understanding, on five external benchmarks
(Omni3D~\citep{brazil2023omni3d}, DA-2K~\citep{yang2024depthanythingv2},
OmniSpatial~\citep{jia2026omnispatial}, MindCube~\citep{wang2025spatial},
and VSI-Bench~\citep{yang2025thinking}) for spatial-reasoning
transfer, and on RoboCasa for manipulation performance.
We further assess geometric generalization on the independently collected
in-house OOD split of Stereo3D-Bench, focusing on point-level correspondence,
disparity, and depth. Evaluation splits are described in \cref{sec:sup_splits}.
Settings specific to each experiment are provided in the corresponding
subsections.

\subsection{Learning Transferable Binocular Representations}
\label{sec:metric_reasoning_analysis}

We analyze how the choice of geometric supervision affects metric-depth
generalization. We compare monocular depth, binocular depth, and
binocular disparity under a controlled geometry-only setting. We then
analyze whether joint disparity--depth training learns the calibrated
conversion between the two quantities and whether geometric supervision
benefits object-level spatial understanding.

\begin{wraptable}{r}{0.51\textwidth}
\vspace{-3.5ex}
\centering
\setlength{\belowcaptionskip}{5pt}
\setlength{\abovecaptionskip}{4pt}
\caption{Depth evaluation on the main and out-of-domain splits of Stereo3D-Bench.
$\dagger$ denotes depth recovered from disparity.}
\label{tab:geometry_generalization}
\fontsize{8}{9.5}\selectfont
\setlength{\tabcolsep}{0pt}
\renewcommand{\arraystretch}{0.98421}
\def\generalizationnum#1{\makebox[2.25em][r]{#1}}
\begin{tabular}{@{}w{l}{0.43\linewidth}*{3}{w{c}{0.19\linewidth}}@{}}
\toprule[0.8pt]
Supervision & $\delta_1$ $\uparrow$ & MAE $\downarrow$ & AbsRel $\downarrow$ \\
\midrule[0.4pt]
\multicolumn{4}{@{}l}{{\fontsize{7.5}{9}\selectfont\itshape Main Split}} \\[1pt]
Monocular depth
& \generalizationnum{77.91} & \generalizationnum{14.08} & \generalizationnum{18.48} \\
Binocular depth
& \generalizationnum{87.26} & \generalizationnum{10.09} & \generalizationnum{10.19} \\
\rowcolor{stereooursbg}
Binocular disparity$^\dagger$
& \generalizationnum{\textbf{91.07}} & \generalizationnum{\textbf{8.68}} & \generalizationnum{\textbf{9.60}} \\
\addlinespace[5pt]
\multicolumn{4}{@{}l}{{\fontsize{7.5}{9}\selectfont\itshape OOD Split}} \\[1pt]
Monocular depth
& \generalizationnum{51.00} & \generalizationnum{26.08} & \generalizationnum{36.75} \\
Binocular depth
& \generalizationnum{45.89} & \generalizationnum{24.64} & \generalizationnum{33.69} \\
\rowcolor{stereooursbg}
Binocular disparity$^\dagger$
& \generalizationnum{\textbf{74.92}} & \generalizationnum{\textbf{18.65}} & \generalizationnum{\textbf{17.88}} \\
\bottomrule[0.8pt]
\end{tabular}
\vspace{-1ex}
\end{wraptable}

\vspace{-1.5ex}
\paragraph{Which geometric supervision generalizes better?}
We initialize all three variants from the same Qwen3-VL-2B-Instruct
checkpoint and train them exclusively on geometry QA constructed from the
same 234K rectified stereo pairs, using an identical optimization budget.
The monocular-depth variant receives
only the left image and directly predicts metric depth. The binocular-depth
variant receives the ordered stereo pair and camera calibration and also
predicts depth directly. The binocular-disparity variant predicts horizontal
disparity from the ordered pair; at evaluation, depth is
recovered using calibration via \cref{eq:disparity_to_depth}.

\Cref{tab:geometry_generalization} shows that the benefit of binocular
input depends on the supervision target. Compared with monocular supervision,
direct binocular depth supervision increases $\delta_1$ from 77.91\% to
87.26\% on the main split, but decreases it from 51.00\% to 45.89\% on
the OOD split. In contrast, binocular disparity supervision
achieves the best results across all three depth metrics on both splits.
On the OOD split, it reaches a $\delta_1$ of 74.92\%, outperforming direct
binocular depth supervision by 29.03 percentage points. These results
suggest that binocular input alone is insufficient for reliable
generalization and support disparity as an intermediate representation
for metric depth estimation across unseen scenes.

\begin{table*}[t]
\vspace{-1ex}
\centering
\setlength{\belowcaptionskip}{5pt}
\setlength{\abovecaptionskip}{4pt}
\sbox0{\parbox[t]{0.49\textwidth}{%
    \caption{Depth estimation via direct prediction and disparity-derived depth (DFD) using the same jointly trained checkpoint at each scale.}%
    \label{tab:depth_consistency}%
}}%
\sbox2{\parbox[t]{0.48\textwidth}{%
    \caption{Effect of geometry QA supervision on object-level performance. Gains are reported in percentage points.}%
    \label{tab:hierarchical_supervision}%
}}%
\dimen0=\dimexpr\ht0+\dp0\relax
\ifdim\dimexpr\ht2+\dp2\relax>\dimen0
    \dimen0=\dimexpr\ht2+\dp2\relax
\fi
\def\stereorowpadding{0pt}
\newcommand{\stereopadrow}{%
    \rule[-\dimexpr\dp\csname @arstrutbox\endcsname+\stereorowpadding/2\relax]{0pt}%
    {\dimexpr\ht\csname @arstrutbox\endcsname+\dp\csname @arstrutbox\endcsname+\stereorowpadding\relax}%
}
\newcommand{\stereodepthtable}{%
\fontsize{7}{8.5}\selectfont
    \setlength{\tabcolsep}{0pt}
    \renewcommand{\arraystretch}{1.1}
    \begin{tabular}{@{}w{l}{0.27\linewidth}w{l}{0.175\linewidth}w{c}{0.13\linewidth}*{2}{w{c}{0.155\linewidth}}w{c}{0.115\linewidth}@{}}
        \toprule[0.8pt]
        Model & Prediction & MAE $\downarrow$ & RMSE $\downarrow$ & AbsRel $\downarrow$ & $\delta_1$ $\uparrow$ \\
        \midrule
        \multirow{2}{*}{\ours-2B} & Direct
        & \textbf{7.52} & \textbf{32.54} & \textbf{6.99} & \textbf{94.51} \\
        & DFD & 7.57 & 32.68 & 7.44 & 94.46 \\
        \midrule[0.3pt]
        \multirow{2}{*}{\ours-8B} & Direct
        & \textbf{6.89} & \textbf{29.93} & \textbf{6.55} & \textbf{95.17} \\
        & DFD & 7.06 & 32.63 & 6.97 & 95.15 \\
        \bottomrule[0.8pt]
    \end{tabular}
}
\newcommand{\stereogeometrytable}{%
\fontsize{7}{8.5}\selectfont
    \setlength{\tabcolsep}{0pt}
    \renewcommand{\arraystretch}{1.1}
    \def\ablationhead##1##2{\shortstack[c]{\strut {\fontsize{7}{8.5}\selectfont ##1}\\[-1pt]{\fontsize{6.5}{7.5}\selectfont\strut ##2}}}
    \begin{tabular}{@{}w{l}{0.245\linewidth}w{c}{0.126\linewidth}w{c}{0.165\linewidth}w{c}{0.12\linewidth}w{c}{0.145\linewidth}w{c}{0.199\linewidth}@{}}
        \toprule[0.8pt]
        \raisebox{5pt}{{\fontsize{7}{8.5}\selectfont Supervision}}
        & \ablationhead{Corr.}{mIoU $\uparrow$}
        & \ablationhead{Disp.}{Acc@2px $\uparrow$}
        & \ablationhead{Depth}{$\delta_1$ $\uparrow$}
        & \ablationhead{3D Box}{mIoU $\uparrow$}
        & \ablationhead{Target}{Joint@0.25 $\uparrow$} \\
        \midrule
        \stereopadrow w/o Geometry & 83.39 & 32.33 & 91.52 & 26.24 & 46.15 \\
        \stereopadrow w/ Geometry & \textbf{86.06} & \textbf{53.92} & \textbf{97.39} & \textbf{36.06} & \textbf{48.72} \\
        \midrule[0.3pt]
        \stereopadrow Gain & +2.67 & +21.59 & +5.87 & +9.82 & +2.57 \\
        \bottomrule[0.8pt]
    \end{tabular}
}
\sbox4{\parbox[t]{0.49\textwidth}{\centering\stereodepthtable}}
\sbox6{\parbox[t]{0.48\textwidth}{\centering\stereogeometrytable}}
\dimen2=\dimexpr\ht4+\dp4-\ht6-\dp6\relax
\ifdim\dimen2>0pt
    \edef\stereorowpadding{\the\dimexpr\dimen2/3\relax}
    \sbox6{\parbox[t]{0.48\textwidth}{\centering\stereogeometrytable}}
\fi
\begin{minipage}[t]{0.49\textwidth}
\vspace{0pt}
\centering
\vtop to \dimen0{\box0\vfil}
\usebox4
\end{minipage}\hfill
\begin{minipage}[t]{0.48\textwidth}
\vspace{0pt}
\centering
\vtop to \dimen0{\box2\vfil}
\usebox6
\end{minipage}
\par\vspace{-10pt}
\end{table*}

\vspace{-1.5ex}
\paragraph{Consistency of direct and disparity-derived depth.}
We next examine whether joint disparity--depth supervision connects the two
output spaces. For each jointly trained model, we compare two depth
estimates at the same query points: directly predicted metric depth and
depth recovered from predicted disparity via the calibrated conversion in
\cref{eq:disparity_to_depth}.

\Cref{tab:depth_consistency} shows that both routes achieve similar MAE and
$\delta_1$ at both model scales. The $\delta_1$ gaps are only 0.05 and 0.02
percentage points for StereoVLM-2B and StereoVLM-8B, respectively, with
corresponding MAE differences of 0.05\,cm and 0.17\,cm. Direct prediction
performs slightly better across all reported metrics, with a more pronounced
difference in 8B RMSE (29.93 vs. 32.63\,cm).
Overall, these results support the link between disparity and calibrated
depth under joint supervision while preserving accurate direct prediction.

\vspace{-1.5ex}
\paragraph{Does geometry supervision improve object-level performance?}
We compare the full StereoVLM-2B model in
\Cref{tab:stereo3d_results} with a variant trained without point-level
geometry QA supervision. \Cref{tab:hierarchical_supervision} shows that
this supervision improves all five object-level metrics.
The largest gains occur in object disparity Acc@2px and 3D-box mIoU,
with improvements of 21.59 and 9.82 percentage points.
Correspondence mIoU, object depth $\delta_1$, and target Joint@0.25 also
improve by 2.67, 5.87, and 2.57 points. These results indicate that
supervision on local stereo geometry benefits language-grounded
object-level reasoning, particularly the estimation of cross-view
displacement and 3D extent.

\subsection{Main Results}
\label{sec:main_results}
\label{sec:valid_binocular_evidence}

We compare StereoVLM with general-purpose multi-image VLMs and monocular
depth-specialized VLMs on Stereo3D-Bench, with results reported in
\Cref{tab:stereo3d_results}. StereoVLM and the multi-image baselines receive
the same ordered stereo pair, whereas monocular depth models receive only
the rectified left image. We follow each baseline's official preprocessing
and inference settings, including its model-specific input resolution.
Before scoring, predicted image coordinates and disparities expressed on
the width-1000 normalized scale are converted to pixels at the benchmark
image resolution. Results are reported only for tasks supported by each model.

\begin{table}[t]
    \centering
    \setlength{\belowcaptionskip}{5pt}
    \setlength{\abovecaptionskip}{4pt}
    \caption{Results on the main split of Stereo3D-Bench. General-purpose VLMs and StereoVLM
        receive the correctly ordered stereo pair, whereas monocular depth
        VLMs receive only the rectified left image. Correspondence and
        disparity MAE are measured in pixels; all other results are
        percentages. “–” denotes an unsupported task. Invalid outputs count as failures for $\delta_1$  and the semantic metrics. Correspondence and disparity MAE exclude invalid predictions. Object-level Avg.
        is the unweighted mean of the five object-level tasks.}
    \label{tab:stereo3d_results}
    \fontsize{7.5}{9}\selectfont
    \setlength{\tabcolsep}{0pt}
    \renewcommand{\arraystretch}{1.03889}
    \def\stereohead#1#2{%
        \shortstack[c]{{\scriptsize\strut #1}\\[-1pt]{\tiny\strut #2}}%
    }
    \def\stereomodel#1#2{%
        \mbox{#1~\citep{#2}}%
    }
    \def\tabnum#1{\makebox[2.25em][r]{#1}}
    \begin{tabular}{@{}w{l}{0.265\linewidth}
        *{3}{w{c}{0.080\linewidth}}
        @{\hspace{0.015\linewidth}}
        *{6}{w{c}{0.080\linewidth}}@{}}
        \toprule[0.8pt]
        \multirow[c]{2}{*}[-\dimexpr(\aboverulesep+\belowrulesep+\cmidrulewidth)/2+3.7pt\relax]{Model}
        & \multicolumn{3}{c}{Point-Level Geometry}
        & \multicolumn{6}{c}{Object-Level Tasks} \\
        \cmidrule(lr){2-4}\cmidrule(l){5-10}
        & \stereohead{Corr.}{MAE $\downarrow$}
        & \stereohead{Disp.}{MAE $\downarrow$}
        & \stereohead{Depth}{$\delta_1$ $\uparrow$}
        & \raisebox{3.7pt}{Avg. $\uparrow$}
        & \stereohead{Corr.}{mIoU $\uparrow$}
        & \stereohead{Disp.}{Acc@2px $\uparrow$}
        & \stereohead{Depth}{$\delta_1$ $\uparrow$}
        & \stereohead{3D Box}{mIoU $\uparrow$}
        & \stereohead{Target}{Joint@0.25 $\uparrow$} \\
        \midrule
        \multicolumn{10}{l}{\itshape General-Purpose VLMs} \\[2pt]
        Qwen3-VL-2B~\citep{bai2025qwen3}
        & \tabnum{73.37}
        & \tabnum{66.92}
        & \tabnum{16.11}
        & \tabnum{7.17}
        & \tabnum{22.89}
        & \tabnum{0.62}
        & \tabnum{12.32}
        & \tabnum{0.00}
        & \tabnum{0.00} \\

        Qwen3-VL-8B~\citep{bai2025qwen3}
        & \tabnum{46.87}
        & \tabnum{41.25}
        & \tabnum{21.47}
        & \tabnum{9.12}
        & \tabnum{24.14}
        & \tabnum{1.23}
        & \tabnum{20.11}
        & \tabnum{0.12}
        & \tabnum{0.00} \\

        GPT-5.6 Sol~\citep{openai2026gpt56sol}
        & \tabnum{28.03}
        & \tabnum{26.36}
        & \tabnum{17.67}
        & \tabnum{9.54}
        & \tabnum{27.83}
        & \tabnum{7.79}
        & \tabnum{11.75}
        & \tabnum{0.34}
        & \tabnum{0.00} \\

        GPT-6 Astra~\citep{openai2026gpt6astra}
        & \tabnum{9.27}
        & \tabnum{7.55}
        & \tabnum{75.96}
        & \tabnum{47.61}
        & \tabnum{68.88}
        & \tabnum{45.05}
        & \tabnum{93.40}
        & \tabnum{17.92}
        & \tabnum{12.82} \\

        \midrule[0.4pt]
        \multicolumn{10}{l}{\itshape Monocular Depth VLMs} \\[2pt]
        \stereomodel{DepthLM-13B}{yeh2026depthlm}
        & --
        & --
        & \tabnum{28.40}
        & --
        & --
        & --
        & --
        & --
        & -- \\

        \stereomodel{DepthVLM-4B}{yu2026unlocking}
        & --
        & --
        & \tabnum{37.11}
        & \tabnum{14.35}
        & \tabnum{20.07}
        & \tabnum{0.69}
        & \tabnum{50.44}
        & \tabnum{0.53}
        & \tabnum{0.00} \\

        \midrule[0.3pt]
        \multicolumn{10}{l}{\itshape Binocular VLMs} \\[2pt]
        \textbf{StereoVLM-2B}
        & \tabnum{4.09}
        & \tabnum{4.12}
        & \tabnum{94.51}
        & \tabnum{64.43}
        & \tabnum{86.06}
        & \tabnum{53.92}
        & \tabnum{97.39}
        & \tabnum{36.06}
        & \tabnum{48.72} \\

        \textbf{StereoVLM-8B}
        & \tabnum{\textbf{3.80}}
        & \tabnum{\textbf{3.76}}
        & \tabnum{\textbf{95.17}}
        & \tabnum{\textbf{66.98}}
        & \tabnum{\textbf{87.62}}
        & \tabnum{\textbf{60.52}}
        & \tabnum{\textbf{97.76}}
        & \tabnum{\textbf{37.70}}
        & \tabnum{\textbf{51.28}} \\
        \bottomrule[0.8pt]
    \end{tabular}
    \vspace{-3.5ex}
\end{table}

\vspace{-1ex}
\paragraph{Point-level geometry.}
StereoVLM-8B achieves the best results across all three point-level tasks
in \Cref{tab:stereo3d_results}. Under the same stereo-input protocol, it
reduces correspondence MAE from 9.27 to 3.80 pixels and disparity MAE from
7.55 to 3.76 pixels relative to GPT-6 Astra, while increasing depth
$\delta_1$ from 75.96\% to 95.17\%. StereoVLM-2B also surpasses GPT-6 Astra
on all three tasks, reaching 94.51\% in depth $\delta_1$. The remaining
general-purpose VLMs exhibit substantially weaker stereo geometric
performance, whereas the depth-specialized baselines receive only the left
image and do not exploit binocular evidence in this evaluation. Compared
with the single-view DepthVLM-4B reference, StereoVLM-8B improves depth
$\delta_1$ by 58.06 percentage points. Together, these results show
consistent improvements in image-space geometry and metric-depth
estimation across both model scales.

\vspace{-1ex}
\paragraph{Language-grounded 3D reasoning.}
Across the five object-level tasks, StereoVLM-2B and StereoVLM-8B achieve
average scores of 64.43\% and 66.98\%, outperforming GPT-6 Astra by 16.82
and 19.37 percentage points, respectively. The advantage is particularly
pronounced in 3D localization: StereoVLM-8B increases oriented 3D-box mIoU
from 17.92\% to 37.70\% and target Joint@0.25 from 12.82\% to 51.28\%.
Both variants also surpass the single-view DepthVLM-4B reference, which
achieves an object-level average of 14.35\%. Scaling from 2B to 8B improves
all five object-level metrics. These results demonstrate stronger 3D
grounding of language-referred objects and task targets, complementing
the gains in point-level geometry. We also outperform cascaded pipelines in accuracy and efficiency
(\cref{sec:sup_modular}).

\begin{figure}[t]
    \vspace{-0ex}
    \centering
    \includegraphics[width=\linewidth]
    {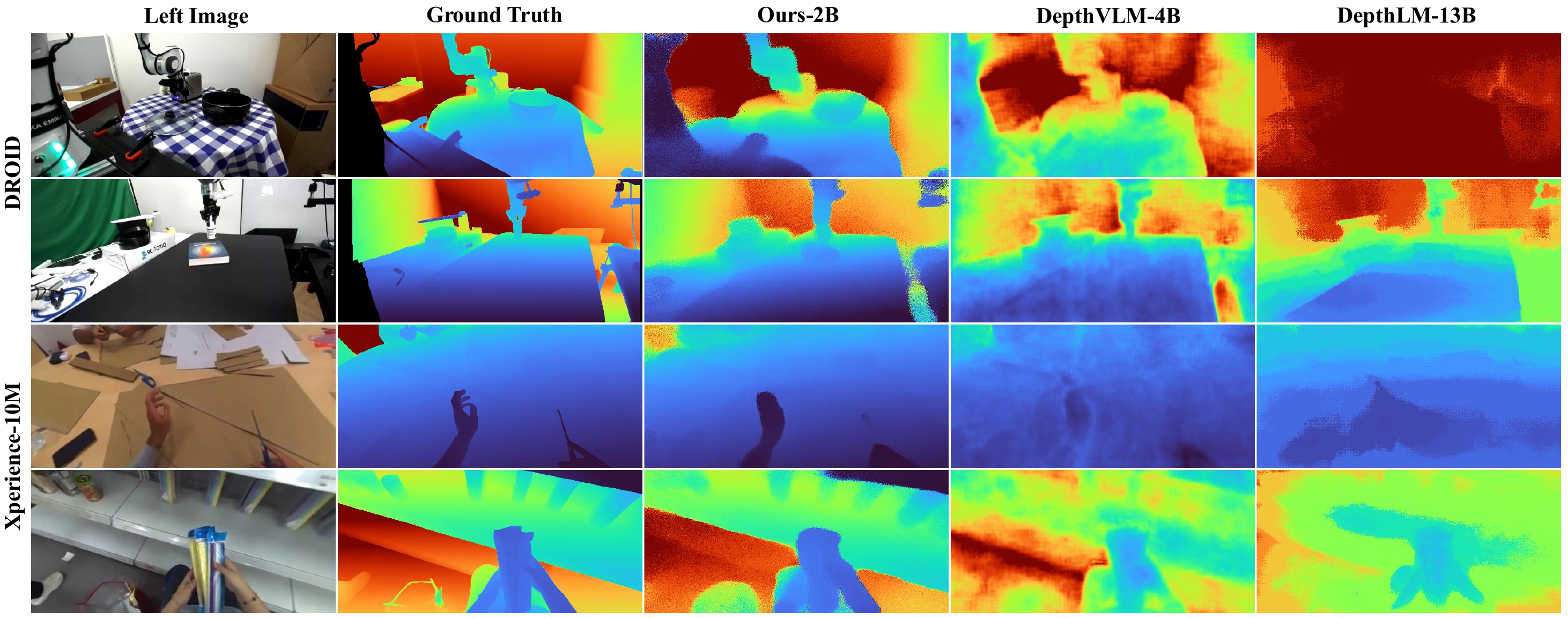}
    \vspace{-5ex}
    \caption{Qualitative comparison of metric-depth predictions on Stereo3D-Bench.}
    \label{fig:depth_vis}
    \vspace{-3ex}
\end{figure}

\vspace{-1ex}
\paragraph{Depth visualization.}
As shown in \cref{fig:depth_vis}, we qualitatively compare StereoVLM-2B with
existing depth-specialized VLMs on representative samples from Stereo3D-Bench.
StereoVLM-2B produces the most faithful depth reconstructions among the compared
methods, recovering the overall scene geometry while better preserving object
boundaries and foreground structure. Notably, StereoVLM-2B achieves these
results with substantially fewer parameters than the two baselines:
approximately half as many as DepthVLM-4B and one-sixth as many as DepthLM-13B.


\begin{wraptable}{R}{0.45\textwidth}
    \vspace{-3ex}
    \centering
    \setlength{\belowcaptionskip}{2pt}
    \setlength{\abovecaptionskip}{2pt}
    \caption{Effect of right-view interventions on StereoVLM-2B.
        Avg. denotes the mean across five object-level tasks.}
    \label{tab:input_interventions}
        \fontsize{7}{8.5}\selectfont
        \setlength{\tabcolsep}{0pt}
        \renewcommand{\arraystretch}{1.1}
        \def\interventionnum#1{\makebox[2.15em][r]{#1}}
        \def\interventionhead#1{{\fontsize{6}{7}\selectfont #1}}
        \begin{tabular}{@{}w{l}{0.40\linewidth}
            w{c}{0.20\linewidth}
            @{\hspace{0.018\linewidth}}
            w{c}{0.167\linewidth}
            w{c}{0.215\linewidth}@{}}
            \toprule[0.8pt]
            \multirow[c]{2}{*}[-\dimexpr(\aboverulesep+\belowrulesep+\cmidrulewidth)/2\relax]{Intervention}
            & \multicolumn{2}{c}{Geometry}
            & Object-Level \\
            \cmidrule(l{2pt}r{2pt}){2-3}\cmidrule(l{2pt}){4-4}
            & \interventionhead{Disp. MAE $\downarrow$}
            & \interventionhead{Depth $\delta_1$ $\uparrow$}
            & \interventionhead{Avg. $\uparrow$} \\
            \midrule
            Duplicated left\,$(\mathbf{I}_l,\mathbf{I}_l)$
            & \interventionnum{32.85}
            & \interventionnum{7.40}
            & \interventionnum{8.18} \\
            Single left\,$\mathbf{I}_l$
            & \interventionnum{9.07}
            & \interventionnum{80.82}
            & \interventionnum{40.26} \\
            Zero-right\,$(\mathbf{I}_l,\mathbf{0})$
            & \interventionnum{9.31}
            & \interventionnum{79.70}
            & \interventionnum{39.81} \\
            Shuffled right\,$(\mathbf{I}_l,\mathbf{I}^{\prime}_r)$
            & \interventionnum{12.22}
            & \interventionnum{70.16}
            & \interventionnum{32.49} \\
            \rowcolor{stereooursbg}
            \textbf{Correct stereo}\,$(\mathbf{I}_l,\mathbf{I}_r)$
            & \interventionnum{\textbf{4.12}}
            & \stereogap{0.018\linewidth}\interventionnum{\textbf{94.51}}
            & \interventionnum{\textbf{64.43}} \\
            \bottomrule[0.8pt]
        \end{tabular}
        \vspace{-3ex}
\end{wraptable}

\paragraph{Binocular evidence and single-view capability.}
To assess whether StereoVLM's gains depend on valid binocular evidence,
we evaluate StereoVLM-2B under the five input conditions in
\Cref{tab:input_interventions}. Only the right-view input changes;
the checkpoint, left image, query, calibration, and decoding settings
remain fixed. Note that the \emph{single-left} control retains stereo
calibration, whereas the single-view protocol in
Appendix~\ref{sec:sup_single_view} also removes the baseline information.
Correct stereo achieves the best results across all reported metrics:
relative to single-left input, disparity MAE decreases from 9.07 to
4.12 pixels, depth $\delta_1$ increases from 80.82\% to 94.51\%, and
the object-level average rises from 40.26\% to 64.43\%. Zero-right
performs similarly to single-left, indicating that retaining a second
image slot without right-view content does not recover the stereo gain.
Duplicated-left and shuffled-right both perform worse than single-left
across all reported metrics, with duplicated-left producing the largest
degradation. This shows that an invalid second view can impair prediction,
rather than simply remove the benefit of stereo input. Together, these
controls support the importance of correctly paired right-view evidence
for geometric and object-level accuracy, while the single-left results
show retained single-view capability. Baseline perturbations further support
the model's use of the supplied baseline for depth prediction
(\cref{sec:sup_baseline_calibration}).

\subsection{Transfer Beyond Stereo3D-Bench}
\label{sec:transfer}

We test spatial reasoning and embodied control using matched 2B models trained with and without stereo instructions. Both share the same initialization, optimization schedule, and training budget. Removing stereo instructions preserves the relative sampling proportions of the remaining categories.

\paragraph{General spatial reasoning.}
We evaluate general spatial reasoning on three monocular benchmarks
(Omni3D, DA-2K, and OmniSpatial) and two video benchmarks
(MindCube and VSI-Bench). \Cref{tab:stereo_data_ablation} shows that adding
stereo instructions improves all five benchmarks, increasing the overall
average from 47.94 to 51.48, a gain of 3.54 percentage points. The largest
gain occurs on Omni3D at 8.21 points, while the video benchmarks MindCube
and VSI-Bench improve by 3.71 and 0.92 points, respectively. With shared spatial supervision, correct stereo and single-left training yield similar averages (50.23 and 50.21). Adding correspondence and disparity supervision to correct stereo raises the average to 51.48 (\cref{sec:sup_spatial_transfer}).

\paragraph{Embodied control.}
We further evaluate whether the learned representation benefits downstream
robotic control. We initialize the same GR00T policy~\citep{bjorck2025gr00t} architecture from
the pretrained Qwen3-VL-2B model and representation models trained with
and without stereo instructions. Each policy is trained for 60k steps using 50 demonstrations per task and
evaluated on 24 RoboCasa~\citep{nasiriany2024robocasa} tasks with 50 rollouts per task.
To isolate the benefit of stereo supervision from exposure to RoboCasa
data during representation training, we exclude RoboCasa from the
representation-training mixture for this embodied-control experiment.

\Cref{tab:downstream_vla} shows that stereo instruction tuning increases
manipulation success from 38.1\% to 42.4\%, a gain of 4.3 percentage
points over general instruction tuning alone. It also exceeds the
pretrained Qwen3-VL-2B initialization at 34.9\% by 7.5 points. All policies
share the same stereo-input setting and downstream training protocol.
Further controls support the benefits of geometric supervision and valid
binocular observations for manipulation (\cref{sec:sup_embodied_control}).

\begin{table*}[t]
\vspace{-4pt}
\centering
\setlength{\belowcaptionskip}{5pt}
\setlength{\abovecaptionskip}{4pt}
\sbox0{\parbox[t]{0.66\textwidth}{%
    \caption{\raggedright Effect of stereo instruction tuning on monocular and video spatial~reasoning. StereoVLM-2B is compared with its variant without stereo tuning; gains are in percentage points.}%
    \label{tab:stereo_data_ablation}%
}}%
\sbox2{\parbox[t]{0.32\textwidth}{%
    \caption{\raggedright Effect of stereo tuning on manipulation success~(\%). All policies use stereo input.}%
    \label{tab:downstream_vla}%
}}%
\dimen0=\dimexpr\ht0+\dp0\relax
\ifdim\dimexpr\ht2+\dp2\relax>\dimen0
    \dimen0=\dimexpr\ht2+\dp2\relax
\fi
\newcommand{\spatialtransfertable}{%
    \fontsize{7}{8.5}\selectfont
    \setlength{\tabcolsep}{0pt}
    \renewcommand{\arraystretch}{1.1}
    \def\ablationhead##1{{\fontsize{6.5}{7.5}\selectfont ##1}}
    \def\ablationnum##1{\makebox[2.25em][r]{##1}}
    \def\ablationgain##1##2{\ablationnum{##1}\hspace{1pt}{\fontsize{6.5}{7.5}\selectfont\color{red!80!black}($\uparrow$\,##2)}}
    \begin{tabular}{@{}w{c}{0.08\linewidth}*{5}{w{c}{0.153\linewidth}}w{c}{0.155\linewidth}@{}}
    \toprule[0.8pt]
    \multirow[c]{2}{*}[-\dimexpr(\aboverulesep+\belowrulesep+\cmidrulewidth)/2\relax]{Stereo}
    & \multicolumn{3}{c}{Monocular}
    & \multicolumn{2}{c}{Video}
    & \multirow[c]{2}{*}[-\dimexpr(\aboverulesep+\belowrulesep+\cmidrulewidth)/2\relax]{Avg. $\uparrow$} \\
    \cmidrule(lr){2-4}\cmidrule(lr){5-6}
    & \ablationhead{Omni3D}
    & \ablationhead{DA-2K}
    & \ablationhead{OmniSpatial}
    & \ablationhead{MindCube}
    & \ablationhead{VSI-Bench}
    & \\
    \midrule[0.4pt]
    --
    & \ablationnum{23.95} & \ablationnum{73.79} & \ablationnum{40.18}
    & \ablationnum{44.86} & \ablationnum{56.94} & \ablationnum{47.94} \\
    \noalign{\begingroup\color{stereooursbg}\hrule height9.35pt\endgroup\kern-9.35pt}
    $\checkmark$
    & \ablationgain{32.16}{8.21} & \ablationgain{76.16}{2.37} & \ablationgain{42.66}{2.48}
    & \ablationgain{48.57}{3.71} & \ablationgain{57.86}{0.92} & \ablationgain{51.48}{3.54} \\
    \bottomrule[0.8pt]
    \end{tabular}
}
\def\vlaheaderpadding{0pt}
\newcommand{\vlatransfertable}{%
    \fontsize{7}{8.5}\selectfont
    \setlength{\tabcolsep}{0pt}
    \renewcommand{\arraystretch}{1.1}
    \def\vlanum##1{\makebox[2.15em][r]{##1}}
    \def\vlahead##1{{\fontsize{6.5}{7.5}\selectfont ##1}}
    \begin{tabular}{@{}w{l}{0.40\linewidth}w{c}{0.21\linewidth}w{c}{0.18\linewidth}w{c}{0.21\linewidth}@{}}
    \toprule[0.8pt]
    \rule[-\dimexpr\dp\csname @arstrutbox\endcsname+\vlaheaderpadding/2\relax]{0pt}%
    {\dimexpr\ht\csname @arstrutbox\endcsname+\dp\csname @arstrutbox\endcsname+\vlaheaderpadding\relax}%
    \vlahead{Model} & \vlahead{General} & \vlahead{Stereo} & \vlahead{Success} \\
    \midrule[0.4pt]
    \multirow{2}{*}{Qwen3-VL-2B} & -- & -- & \vlanum{34.9} \\
    & $\checkmark$ & -- & \vlanum{38.1} \\
    \rowcolor{stereooursbg}
    \textbf{StereoVLM-2B} & $\checkmark$ & $\checkmark$ & \vlanum{\textbf{42.4}} \\
    \bottomrule[0.8pt]
    \end{tabular}
}
\sbox4{\parbox[t]{0.66\textwidth}{\centering\spatialtransfertable}}
\sbox6{\parbox[t]{0.32\textwidth}{\centering\vlatransfertable}}
\dimen2=\dimexpr\ht4+\dp4-\ht6-\dp6\relax
\ifdim\dimen2>0pt
    \edef\vlaheaderpadding{\the\dimen2}
    \sbox6{\parbox[t]{0.32\textwidth}{\centering\vlatransfertable}}
\fi
\begin{minipage}[t]{0.66\textwidth}
\vspace{0pt}
\centering
\vtop to \dimen0{\box0\vfil}
\usebox4
\end{minipage}\hfill%
\begin{minipage}[t]{0.32\textwidth}
\vspace{0pt}
\centering
\vtop to \dimen0{\box2\vfil}
\usebox6
\end{minipage}
\par\vspace{-14pt}
\end{table*}

%% file: sec/6_conclusion.tex
\section{Conclusion and Limitations}
\label{sec:conclusion}
We present StereoVLM, a geometry-aware binocular vision-language model for metric 3D spatial reasoning, trained with hierarchical stereo supervision from cross-view correspondence and disparity to calibration-conditioned metric depth and semantic spatial reasoning. We show that disparity supervision provides substantially stronger metric-depth generalization to unseen environments than direct binocular depth supervision, while controlled interventions verify that StereoVLM benefits from valid cross-view geometry. Hierarchical geometric supervision further benefits broader spatial reasoning and downstream embodied control. The current study is limited by its focus on controlled benchmarks and simulated kitchen tasks, as well as its reliance on calibrated and synchronized cameras; future work should address the simulation-to-real gap and improve robustness to challenging visual conditions and diverse robot platforms.

%% file: appendix/detail.tex
\appendix
\newpage

\section*{Supplementary Material}
This supplementary material documents the construction and evaluation of
Stereo3D-Bench, together with additional training details and experiments
for \ours. Training and benchmark pairs follow a shared processing and
annotation procedure but are kept separate by the data split. The sections
below describe this procedure, the split, the evaluation protocol, and
supporting results.\\
\noindent $\triangleright$ \textbf{\cref{sec:sup_data}}: Geometric
references, semantic annotations, and query and answer schemas, with
point-level and object-level examples.\\
\noindent $\triangleright$ \textbf{\cref{sec:sup_splits}}: The training
mixture, the main and OOD splits of Stereo3D-Bench, and per-source counts
for the main split.\\
\noindent $\triangleright$ \textbf{\cref{sec:sup_protocol}}: Metric
definitions, invalid outputs, and the implementation of right-view
interventions.\\
\noindent $\triangleright$ \textbf{\cref{sec:sup_results}}: Training
settings, geometric generalization, input controls, extrinsic robustness,
baseline sensitivity, transfer experiments, and modular stereo--VLM comparisons.
\vspace{-5pt}

\section{Data Construction}
\label{sec:sup_data}

Training data and Stereo3D-Bench share the same stereo representation,
question schemas, and coordinate conventions, but intentionally use
different geometric-reference pathways. Training supervision uses FoundationStereo predictions, whereas
Stereo3D-Bench is scored against source-native geometric references. This separation prevents benchmark
performance from directly measuring reproduction of the stereo estimator
used to construct the training data. The following subsections describe
the shared representation, the training and evaluation reference pathways,
the semantic annotations, and question generation. Dataset splitting and
leakage control are described in \cref{sec:sup_splits}.

\subsection{Unified Stereo Data Representation}
\label{sec:sup_representation}

Every source sample is converted into an ordered, rectified stereo pair at
a resolution of $800\times320$. During preprocessing, camera intrinsics
are standardized across calibrated stereo datasets. We retain the rectified camera metadata,
including the horizontal focal length $f_x$, stereo baseline $B$, and
camera extrinsics when provided by the source. Disparity is expressed in
rectified-left pixel coordinates, and physical depth is represented by the
left-camera optical-axis distance in meters.

\begin{figure*}[t]
  \centering
  \includegraphics[width=\textwidth]
  {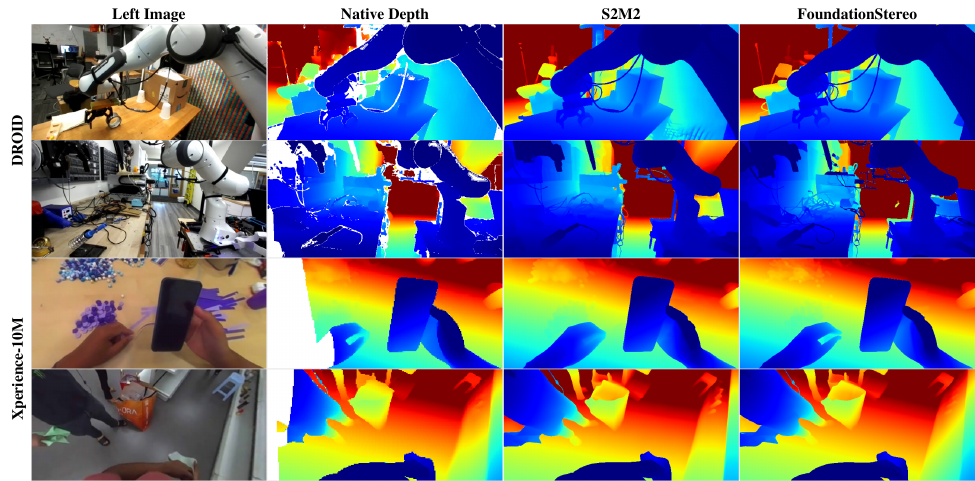}
  \vspace{-4ex}
  \caption{Comparison of source-native and estimated left-view depth.
  The first two rows show DROID examples and the last two show
  Xperience-10M examples. Columns show the rectified left image, the
  source-native depth, S2M2 depth, and FoundationStereo depth. Predicted
  depth is converted from disparity using the physical focal length and
  baseline released for each pair. All maps are aligned to the
  rectified-left image; colors are normalized independently within each
  row and should not be compared across rows. White denotes invalid depth in the corresponding map. Rectification borders are cropped for display.}
  \label{fig:sup_representation}
\end{figure*}

Training and evaluation pairs use the same representation but differ in
reference provenance. FoundationStereo~\citep{wen2025foundationstereo}
provides dense disparity supervision for the training mixture. In
Stereo3D-Bench, DROID uses its aligned sensor depth, Xperience-10M uses its
officially released stereo-depth annotation, RoboCasa uses simulator
geometry, and FSD uses its source-native disparity reference. Because FSD
does not provide physical metric calibration, it contributes
correspondence and disparity questions but not metric-depth questions.
\Cref{fig:sup_representation} compares the source-native references with
the outputs of S2M2 and FoundationStereo.

\subsection{Point-Level Geometric References}
\label{sec:sup_geometry}

\paragraph{Training references.}
For training samples with source-native geometric references, we retain
pixels whose native reference is valid and agrees with FoundationStereo.
FoundationStereo then provides the numerical disparity target on these
filtered regions. When physical calibration is available, disparity is
converted to metric depth as
\begin{equation}
  Z = \frac{f_xB}{d}.
\end{equation}
Thus, the training target is provided by FoundationStereo, while
source-native references are used to identify reliable training regions.

\paragraph{Benchmark references.}
Stereo3D-Bench uses source-native geometry rather than FoundationStereo
outputs. For DROID, the numerical depth target is the aligned ZED sensor
depth. For Xperience-10M, it is the officially released metric stereo-depth
annotation. For RoboCasa, depth and disparity are obtained directly from
the simulator. FSD supplies source-native disparity but no physical
metric-depth reference and is therefore excluded from metric-depth
evaluation. When a physical depth reference $Z$ is available, its disparity
and right-image correspondence are derived using
\begin{equation}
  d = \frac{f_xB}{Z},
  \qquad
  x^r = x^l-d.
\end{equation}

For the two real-image sources, DROID and Xperience-10M, we use
S2M2~\citep{min2025s}, a separately trained stereo estimator that is
not used to generate our training targets, as a reliability estimator. After
converting its disparity to depth using the same physical calibration, a
pixel is retained when both depths are valid and
\begin{equation}
  \frac{\left|Z_{\mathrm{S2M2}}-Z_{\mathrm{native}}\right|}
       {Z_{\mathrm{native}}}
  \leq 0.10.
\end{equation}
S2M2 determines only the validity mask; all benchmark answers are still
computed from the source-native reference. Importantly, FoundationStereo,
which supplies the training supervision, is not used to select benchmark
pixels or to define their answers.

We additionally remove pixels within 16 pixels of the image boundary and
pixels whose projected right-image locations fall outside the common valid
crop. Local neighborhood checks suppress isolated valid pixels and
discontinuities unsupported by neighboring reference values. These checks
are applied before any evaluated model is run, and sampling uses a fixed
random seed.

\paragraph{Point sampling.}
Retained pixels are divided into a disparity-edge band and the remaining
valid region. The edge band contains retained pixels within three pixels
of a ridge in the source-native disparity map, ensuring that the candidate
pool includes geometric boundaries rather than only large smooth surfaces.
In the fixed Stereo3D-Bench protocol, each evaluated family contains 120
sampled points per pair, grouped into 15 questions of eight points each.
Correspondence, disparity, and depth use separately sampled points but the
same source-native reference and coordinate convention.

\begin{figure*}[t]
  \centering
  \includegraphics[width=\textwidth]
  {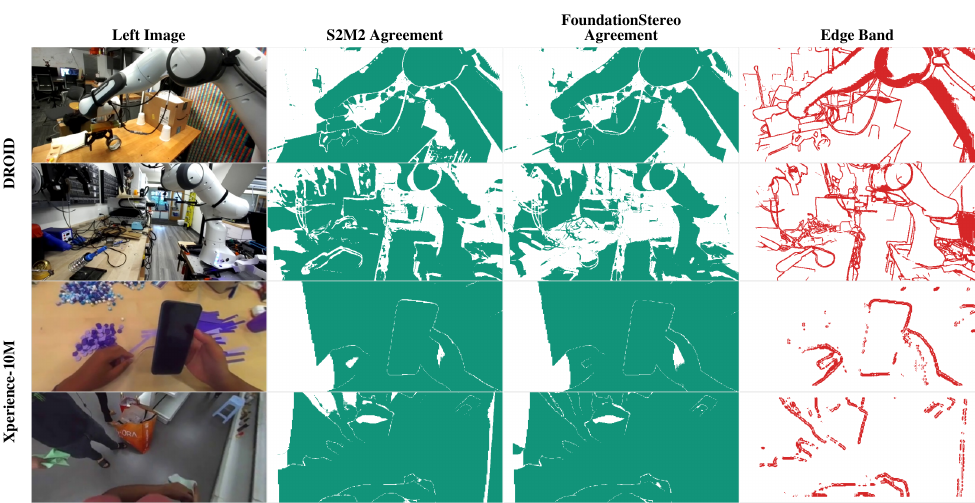}
  \vspace{-4ex}
  \caption{Point-level reference filtering. The first two rows show DROID
  and the last two show Xperience-10M. Green pixels in the second column
  have S2M2 depth within 10\% of the source-native depth and form the
  benchmark validity mask. The third column shows the analogous
  FoundationStereo agreement mask for comparison only; it is not used for
  benchmark sampling. Red pixels in the final column lie within three
  pixels of a ridge in the source-native disparity map. Rectification
  borders are cropped for display.}
  \label{fig:sup_geometry}
\end{figure*}

\Cref{fig:sup_geometry} visualizes the reliability masks. The
FoundationStereo agreement mask is included only as a diagnostic comparison
with the training teacher; it is not used to construct Stereo3D-Bench.

\subsection{Object-Level Semantic References}
\label{sec:sup_semantic}

Object references are built from the rectified left image.
Qwen3~\citep{bai2025qwen3} proposes object names,
SAM3~\citep{carion2026sam} produces their masks, and
DAM~\citep{lian2025describe} supplies region descriptions used to reject
semantically inconsistent proposals. The right image is not segmented.
For benchmark pairs, object disparity, depth, right-image location, and
lifted 3D geometry use the native references in \cref{sec:sup_geometry},
rather than FoundationStereo estimates.

Valid object pixels are lifted into the left-camera point cloud and
expressed in a gravity-aligned frame, where object extent and yaw are
estimated. The resulting oriented 3D box is then transformed back to the
rectified-left camera frame. DROID reaches the gravity-aligned frame using
the provided camera extrinsics. For the egocentric Xperience-10M images,
the gravity-aligned frame is estimated because the camera optical axis is
not generally perpendicular to the ground.

\begin{figure*}[t]
  \centering
  \includegraphics[width=\textwidth]{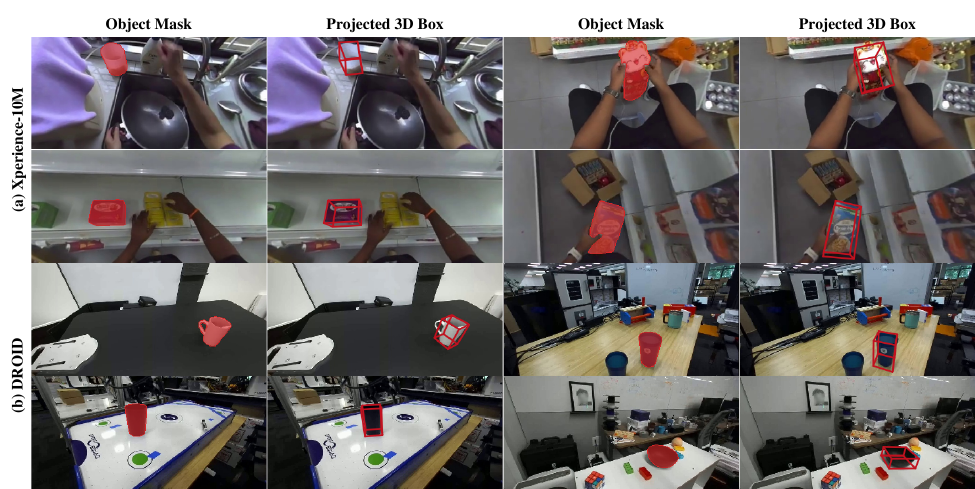}
  \vspace{-4ex}
  \caption{Object references.
  (a)~Xperience-10M, using the estimated gravity-aligned frame.
  (b)~DROID, using the provided camera extrinsics.
  In each pair, the left image shows the object mask and the right image projects the oriented box after it is written back to the left camera.}
  \label{fig:sup_semantic}
\end{figure*}

These references instantiate the five object-level tasks defined in
\cref{sec:benchmark_tasks}: object correspondence, object disparity,
object depth, oriented 3D grounding, and task-target grounding.
See \cref{fig:sup_semantic}.

\subsection{Question Generation and Quality Control}
\label{sec:sup_questions}

A question is formed from the references above and is posed on the ordered stereo pair. A point-level question lists one or more pixels in the left image, and the answer returns one value per pixel in the same order. Correspondence is a right-image coordinate, disparity is expressed on a normalized scale whose image width is 1000, and depth is in meters. The focal length and baseline are written into depth questions. An object-level question refers to an object by language. It asks for the object's right-image box, its disparity, its depth, its
oriented 3D box in the left-camera frame, or the label and 3D box of the
object to manipulate.

Both training and Stereo3D-Bench apply the same automatic filters. Part of the proposed object list is removed. A 3D box is kept only when the box projected back onto the image has sufficient overlap with the object mask. Stereo3D-Bench then receives a manual review, and questions that remain low quality are removed. Training questions do not receive this review.

\subsection{Question and Answer Schemas}
\label{sec:sup_schemas}

We generate questions from task-specific templates. Each task has a fixed
answer schema and multiple question paraphrases: two each for point
correspondence, disparity, and depth; three each for object correspondence,
disparity, and depth; five for oriented 3D grounding; and three for
task-target grounding. \Cref{tab:sup_prompt_examples} shows one
paraphrase per task. Placeholders are replaced with the
sample's coordinates, object reference, instruction, and calibration.

Point queries list one or more pixels in the left image. For multiple
pixels, the answer contains one numbered line per pixel in the same order;
a single-pixel query returns only the requested value. Disparities are
reported in width-normalized coordinates,
$d_{1000}=1000d_{\mathrm{px}}/W$, where $W$ is the image width in pixels.
Depth is the left-camera $z$ coordinate in meters. Metric-depth questions
state the rectified focal length and stereo baseline.

Object correspondence returns a right-image box
$[x_1,y_1,x_2,y_2]$. Object disparity and depth return a single numeric
value in the same units as their point-level counterparts. Oriented 3D
grounding and task-target grounding return an object label and a 3D box
expressed in the left-camera frame.

\begin{table*}[t]
\centering
\setlength{\abovecaptionskip}{4pt}
\setlength{\belowcaptionskip}{5pt}
\caption{Representative question templates; angle brackets denote sample-specific fields.}
\label{tab:sup_prompt_examples}
\fontsize{7.5}{9}\selectfont
\setlength{\tabcolsep}{0pt}
\renewcommand{\arraystretch}{1.03889}
\begin{tabular}{@{}p{0.20\linewidth}@{\hspace{0.02\linewidth}}p{0.78\linewidth}@{}}
\toprule[0.8pt]
Task & Question Template \\
\midrule
\multicolumn{2}{@{}l}{\textit{Point-level geometry}} \\
\addlinespace[2pt]
Correspondence &
For each of the following pixels in the left stereo image, locate the
projection of the same point in the right stereo image.
Pixels: \texttt{<numbered pixels>}.
Answer one line per pixel as \texttt{1. (x, y)}, in the same order. \\
\addlinespace[3pt]
Disparity &
For each of the following pixels in the left stereo image, give the
horizontal disparity to its corresponding pixel in the right stereo image.
Express each disparity in the normalized scale where the image width equals
1000. Pixels: \texttt{<numbered pixels>}.
Answer one line per pixel as \texttt{1. <value>}, in the same order. \\
\addlinespace[3pt]
Depth &
For each of the following pixels in the left stereo image, estimate its
depth along the camera $z$-axis using rectified stereo geometry.
The stereo pair is rectified. The focal length is \texttt{<fx>} pixels
and the baseline is \texttt{<B>} meters.
Pixels: \texttt{<numbered pixels>}.
Answer one line per pixel as \texttt{1. <value>}, using meters,
in the same order. \\
\midrule
\multicolumn{2}{@{}l}{\textit{Object-level semantics}} \\
\addlinespace[2pt]
Correspondence &
In the left stereo image, identify \texttt{<object>}. Then locate the
same physical object in the right stereo image and return its bounding
box. Answer with the right-image bounding box only. \\
\addlinespace[3pt]
Disparity &
In the left stereo image, identify \texttt{<object>}. What is its
horizontal disparity in the right stereo image? Express the disparity
in the normalized scale where the image width equals 1000.
Answer with the numeric value only. \\
\addlinespace[3pt]
Depth &
Stereo camera calibration for this displayed pair:
\texttt{fx=<fx>} pixels, \texttt{baseline=<B>} meters.
What is the $z$-depth of \texttt{<object>} in the left stereo image
along the left camera $z$-axis? Answer with the numeric depth value
in meters only. \\
\addlinespace[3pt]
3D grounding &
Rectified-left camera calibration: \texttt{fx=<fx>},
\texttt{fy=<fy>} pixels; \texttt{baseline=<B>} meters.
The principal point is at the image center.
Predict the 3D oriented bounding box of \texttt{<3d\_ref>} in the
left stereo image. Answer with the 3D bounding box and object label only. \\
\addlinespace[3pt]
Task target &
Rectified-left camera calibration: \texttt{fx=<fx>},
\texttt{fy=<fy>} pixels; \texttt{baseline=<B>} meters.
The principal point is at the image center.
For the task below, identify the visible object that should be
manipulated and predict its 3D oriented bounding box in the left stereo
image. Answer with the 3D bounding box and target label only.
Task: \texttt{<instruction>}. \\
\bottomrule[0.8pt]
\end{tabular}
\end{table*}

\section{Splits and Leakage Control}
\label{sec:sup_splits}

Training data and Stereo3D-Bench follow the data-construction procedure in
\cref{sec:sup_data}, while their source data are separated before stereo
pairs, geometric references, semantic annotations, or questions are
generated. Consequently, no benchmark image or question derived from it is
used during training. \Cref{tab:sup_splits} summarizes the resulting
main-split composition.

For DROID, splitting is performed at the scene level. We define a scene by
the laboratory, robot, and collection date. All recordings, episodes,
camera views, and frames associated with one scene are assigned to the same
split. For Xperience-10M, the recording UUID is used as the scene-grouping
unit, and all episodes and frames from one recording remain together.
Therefore, benchmark scenes from either source contribute neither images
nor derived questions to training.

The remaining sources use source-specific exclusion units. FSD benchmark
images are excluded from training at the source-frame level. RoboCasa
contains 24 official task identities, five of which are reserved for
evaluation. Samples from these five tasks are excluded from the training
mixture, making the corresponding evaluation split disjoint at the
task-identity level. FSD supplies correspondence and disparity references
but no physical metric-depth reference; metric depth is therefore evaluated
on the other 1,517 geometry pairs. The primary semantic benchmark excludes RoboCasa.

The OOD split of Stereo3D-Bench uses an independently collected in-house
dataset and covers only point-level correspondence, disparity, and depth,
with the same task definitions and metrics as the main split. Results are
reported separately in \cref{tab:sup_ood}; \cref{tab:sup_splits} counts
only main-split samples.

In the fixed Stereo3D-Bench geometry protocol, each evaluated family uses
15 grouped questions per pair. DROID, Xperience-10M, and RoboCasa contain
three geometry families, whereas FSD contains correspondence and disparity
only. The geometry QA counts in \cref{tab:sup_splits} refer to the
standard stereo input and do not count repeated evaluations under right-view
interventions.

\begin{table}[t]
\centering
\setlength{\abovecaptionskip}{4pt}
\setlength{\belowcaptionskip}{5pt}
\caption{Composition of the main split of Stereo3D-Bench by source. Geometry and semantic
pairs are selected independently and therefore should not be summed to
obtain a unique-image count.}
\label{tab:sup_splits}
\fontsize{8.5}{10}\selectfont
\setlength{\tabcolsep}{0pt}
\renewcommand{\arraystretch}{0.935}
\def\splitnum#1{#1}
\begin{tabular}{@{}w{l}{0.40\linewidth}*{4}{w{c}{0.15\linewidth}}@{}}
\toprule[0.8pt]
\multirow[c]{2}{*}[-\dimexpr(\aboverulesep+\belowrulesep+\cmidrulewidth)/2\relax]{Source}
& \multicolumn{2}{c}{Geometry}
& \multicolumn{2}{c}{Semantic} \\
\cmidrule(lr){2-3}\cmidrule(lr){4-5}
& {Pairs} & {QA} & {Pairs} & {QA} \\
\midrule
DROID~\citep{khazatsky2024droid} & \splitnum{517} & \splitnum{23,265} & \splitnum{556}   & \splitnum{3,833} \\
Xperience-10M~\citep{xperience_10m} & \splitnum{500} & \splitnum{22,500} & \splitnum{1,000} & \splitnum{7,448} \\
FSD~\citep{wen2025foundationstereo} & \splitnum{500} & \splitnum{15,000} & \splitnum{--}    & \splitnum{--} \\
RoboCasa~\citep{nasiriany2024robocasa} & \splitnum{500} & \splitnum{22,500} & \splitnum{--}    & \splitnum{--} \\
\midrule
\textbf{Total}
& \splitnum{\textbf{2,017}}
& \splitnum{\textbf{83,265}}
& \splitnum{\textbf{1,556}}
& \splitnum{\textbf{11,281}} \\
\bottomrule[0.8pt]
\end{tabular}
\end{table}

\section{Evaluation Protocol}
\label{sec:sup_protocol}

\paragraph{Point-level geometry.}
Coordinates and disparities expressed on the width-1000 normalized scale
are converted to pixels at the displayed image resolution before scoring.
In particular, $d_{\mathrm{px}}=Wd_{1000}/1000$ for image width $W$.
Let $\mathcal{P}$ contain all queried points for a task, and let
$\mathcal{P}_{\mathrm{valid}}$ contain those with a successfully parsed,
finite prediction. The primary correspondence error is the horizontal
error of the disparity implied by the predicted right-image coordinate:
\begin{equation}
\mathrm{MAE}_{\mathrm{corr}} =
\frac{1}{|\mathcal{P}_{\mathrm{valid}}|}
\sum_{i\in\mathcal{P}_{\mathrm{valid}}}
\left|(x_i^l-\hat{x}_i^r)-d_i\right|.
\end{equation}
Here $x_i^l$ is the queried left-image coordinate, $\hat{x}_i^r$ is
the predicted right-image coordinate, and $d_i$ is the reference
disparity, all in pixels. The two-dimensional endpoint error and
vertical coordinate error are retained as diagnostic metrics but do
not enter the primary correspondence score. Disparity MAE is
\begin{equation}
\mathrm{MAE}_{d} =
\frac{1}{|\mathcal{P}_{\mathrm{valid}}|}
\sum_{i\in\mathcal{P}_{\mathrm{valid}}}
|\hat d_i-d_i|,
\end{equation}
also in pixels. The valid-point set is determined separately for each
task; therefore, the correspondence and disparity denominators need
not match.

For metric depth, we report MAE in centimeters, AbsRel, and $\delta_1$:
\begin{align}
\mathrm{MAE}_{Z} &=
\frac{1}{|\mathcal{P}_{\mathrm{valid}}|}
\sum_{i\in\mathcal{P}_{\mathrm{valid}}}
|\hat Z_i-Z_i|, \\
\mathrm{AbsRel} &=
\frac{1}{|\mathcal{P}_{\mathrm{valid}}|}
\sum_{i\in\mathcal{P}_{\mathrm{valid}}}
\frac{|\hat Z_i-Z_i|}{Z_i}, \\
\delta_1 &=
\frac{1}{|\mathcal{P}|}
\sum_{i\in\mathcal{P}_{\mathrm{valid}}}
\mathbf{1}\!\left[
\max\!\left(
\frac{\hat Z_i}{Z_i},
\frac{Z_i}{\hat Z_i}
\right)<1.25
\right].
\end{align}
A valid depth prediction must additionally be positive. AbsRel is
multiplied by 100 when reported as a percentage. MAE and AbsRel are
computed over valid predictions, whereas the $\delta_1$ denominator
contains all queried points; invalid depth predictions therefore count
as failures under $\delta_1$. Point-level metrics weight queried points
equally rather than first averaging within each question.

\paragraph{Object-level semantics.}
Object correspondence is measured by the IoU between the predicted and
reference right-image boxes, averaged over all correspondence questions.
An invalid box contributes an IoU of zero. Object disparity uses
$\mathrm{Acc}@2\mathrm{px}$, the fraction of questions satisfying
$|\hat d-d|\leq2$ pixels after conversion from the width-1000 normalized
scale. Object depth uses the $\delta_1$ definition above. Oriented 3D
grounding uses the volumetric IoU between the predicted and reference
oriented boxes in the rectified-left camera frame, averaged over all
questions, with invalid boxes assigned zero IoU. Task-target grounding
is correct only when the predicted target label is correct and its
oriented 3D-box IoU is at least $0.25$. The semantic average is the
unweighted mean of these five task scores after each is expressed as
a percentage.

\paragraph{Invalid outputs.}
A prediction is invalid if it is missing, cannot be parsed into the
required answer schema, contains non-finite values, gives a non-positive
depth, or specifies an invalid 3D box. For a multi-point answer, parsed
entries are scored individually and missing entries are marked invalid;
entries beyond the requested number are ignored. Invalid semantic
predictions receive zero IoU or count as failures for
$\mathrm{Acc}@2\mathrm{px}$, $\delta_1$, and task-target grounding.
Invalid point-level predictions are excluded from MAE and AbsRel but
remain in the coverage denominator; invalid depth predictions also
remain in the $\delta_1$ denominator.

\paragraph{Controlled right-view inputs.}
For each query, the checkpoint, left image, question, calibration, and
decoding configuration are held fixed. Correct stereo $(I_l,I_r)$
provides the primary benchmark score. Four diagnostic conditions alter
the availability or validity of the right view. Single-left input
removes the second image, $(I_l)$. Duplicated-left input places the left
image in both slots, $(I_l,I_l)$. Zero-right input replaces the right
view with an equally sized black image, $(I_l,\mathbf{0})$.
Shuffled-right input replaces the correct right image with a right image
from another benchmark example, $(I_l,I_r')$. Scores under these four
conditions are reported separately and do not enter the primary
Stereo3D-Bench score.

\section{Additional Results}
\label{sec:sup_results}

This section provides the training-data composition and optimization
settings omitted from the main paper, followed by geometric generalization evaluation
and additional analyses of the single-view capability retained after stereo training, robustness to
extrinsic calibration errors, sensitivity to the supplied baseline, and
transfer to spatial reasoning and embodied control. We also provide
qualitative results and compare modular stereo--VLM pipelines.

\subsection{Training Data and Optimization}
\label{sec:sup_training}

\paragraph{Stereo training data.}
We construct the stereo instruction data from DROID, FSD,
Xperience-10M, and RoboCasa. RoboCasa contributes point-level geometry
instructions. All stereo pairs are processed using the construction
procedure in \cref{sec:sup_data}.

\Cref{tab:sup_training_inventory} reports the training inventory after
filtering and data expansion. Geometry and semantic pairs are counted
separately because the two annotation pipelines retain different
subsets of the source images. Geometry QA counts denote complete
question--answer records rather than individual points within a
multi-point question. Semantic counts cover the five object-level task
families used in the final model and benchmark.

\paragraph{Geometry-only training data.}
The three geometry-only variants compared in
\cref{tab:geometry_generalization} are trained on the same 234K rectified
stereo pairs from Xperience-10M, DROID, RoboCasa, and an in-house training
collection.

\paragraph{Instruction mixture.}
Stereo instructions account for \(25\%\) of the final instruction
mixture. We allocate \(15\%\) to point-level stereo geometry and
\(10\%\) to object-level stereo semantics, preserving a \(60{:}40\)
ratio within the stereo portion. The remaining \(75\%\) is distributed
among the monocular and video instruction categories in proportion to
their previous weights. The resulting top-level sampling probabilities
are shown in \cref{fig:sup_training_mix}. These probabilities specify
the composition of the instruction stream after construction of the
stereo inventory above.

\begin{table}[!h]
\centering
\begin{minipage}[t]{0.45\linewidth}
\vspace{0pt}
\centering
\includegraphics[width=\linewidth]{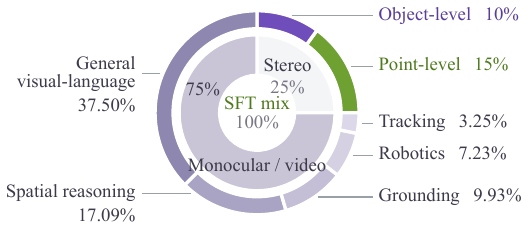}\par
\expandafter\def\csname @captype\endcsname{figure}
\setlength{\abovecaptionskip}{3pt}
\setlength{\belowcaptionskip}{0pt}
\caption{Instruction mixture. The inner ring shows group sampling
proportions; the outer ring shows their constituent categories.}
\label{fig:sup_training_mix}
\end{minipage}\hfill
\begin{minipage}[t]{0.53\linewidth}
\vspace{0pt}
\centering
\expandafter\def\csname @captype\endcsname{table}
\setlength{\abovecaptionskip}{0pt}
\setlength{\belowcaptionskip}{4pt}
\caption{Stereo training inventory. Rounded pair and QA counts are
reported by source for geometry and semantic annotations.}
\label{tab:sup_training_inventory}
\par
\fontsize{8}{9.5}\selectfont
\setlength{\tabcolsep}{1pt}
\renewcommand{\arraystretch}{0.98421}
\begin{tabular*}{\linewidth}{@{\extracolsep{\fill}}l*{4}{c}@{}}
\toprule[0.8pt]
\multirow[c]{2}{*}[-\dimexpr(\aboverulesep+\belowrulesep+\cmidrulewidth)/2\relax]{Source}
& \multicolumn{2}{c}{Geometry}
& \multicolumn{2}{c}{Semantic} \\
\cmidrule(lr){2-3}\cmidrule(lr){4-5}
& Pairs & QA & Pairs & QA \\
\midrule
DROID~\citep{khazatsky2024droid} & 100.0K & 8.40M & 85.2K & 710K \\
FSD~\citep{wen2025foundationstereo} & 1.00M & 84.0M & -- & -- \\
Xperience-10M~\citep{xperience_10m} & 410.3K & 34.5M & 83.6K & 492K \\
RoboCasa~\citep{nasiriany2024robocasa} & 10.0K & 0.84M & -- & -- \\
\midrule
\textbf{Total} & \textbf{1.52M} & \textbf{127.7M} & \textbf{168.8K} & \textbf{1.20M} \\
\bottomrule[0.8pt]
\end{tabular*}
\end{minipage}
\end{table}

\paragraph{Optimization.}
We initialize each StereoVLM variant from the corresponding
Qwen3-VL-Instruct checkpoint and perform full-parameter instruction
tuning, including the vision encoder, language model, and multimodal
aligner. The final 2B and 8B models share the same training data mixture and
optimization schedule. Both are trained for 3,000 optimization steps with an initial
learning rate of \(1\times10^{-5}\), a warmup ratio of \(0.03\), and an
effective global batch size of 64. We use a maximum sequence length of
32,768 tokens, bfloat16 computation, gradient checkpointing, and
DeepSpeed ZeRO-2. Stereo images are presented at \(800\times320\), the
same resolution used to construct the geometric and semantic
references. Dataset and training-order shuffling use a fixed random
seed. The reduced-supervision input controls are described separately in
\cref{sec:sup_transfer}.

\newpage
\suppressfloats[t]
\subsection{Geometric Generalization to Unseen Scenes}
\label{sec:sup_ood}

We further evaluate geometric generalization on the in-house OOD split
of Stereo3D-Bench. \Cref{tab:sup_ood} reports point-level correspondence,
disparity, and depth results using the same metrics as the main split.

\subsection{Single-View Capability and Input Controls}
\label{sec:sup_single_view}

\begin{wraptable}{r}{0.52\textwidth}
\vspace{\dimexpr-3ex-3.02pt\relax}
\centering
\hypersetup{hidelinks}
\setlength{\belowcaptionskip}{5pt}
\setlength{\abovecaptionskip}{4pt}
\caption{Point-level geometry on the OOD split of Stereo3D-Bench.
MAE is in pixels; $\delta_1$ is in percent. Invalid depth outputs count as failures.}
\label{tab:sup_ood}
\fontsize{8.5}{10.5}\selectfont
\setlength{\tabcolsep}{0pt}
\renewcommand{\arraystretch}{0.890476}
\def\oodcite#1{{\fontsize{7}{8}\selectfont\citep{#1}}}
\def\oodhead#1#2{\shortstack[c]{\strut\textbf{#1}\\[-1pt]{\fontsize{7.5}{9}\selectfont\strut #2}}}
\def\oodgroup#1{\multicolumn{4}{@{}l@{}}{{\fontsize{7.5}{9}\selectfont\color{black!65}\itshape #1}}}
\begin{tabular}{@{}w{l}{0.53\linewidth}@{\hspace{0.035\linewidth}}*{3}{w{r}{0.14\linewidth}}@{\hspace{0.015\linewidth}}}
\toprule[0.8pt]
\raisebox{5.5pt}{\textbf{Model}}
& \multicolumn{1}{w{c}{0.14\linewidth}}{\oodhead{Corr.}{MAE $\downarrow$}}
& \multicolumn{1}{w{c}{0.14\linewidth}}{\oodhead{Disp.}{MAE $\downarrow$}}
& \multicolumn{1}{w{c}{0.14\linewidth}}{\oodhead{Depth}{$\delta_1$ $\uparrow$}} \\
\midrule[0.4pt]
\oodgroup{General-Purpose VLMs} \\[1pt]
Qwen3-VL-2B~\oodcite{bai2025qwen3}
& 70.96 & 34.26 & 16.69 \\
Qwen3-VL-8B~\oodcite{bai2025qwen3}
& 32.66 & 27.93 & 26.68 \\
GPT-5.6 Sol~\oodcite{openai2026gpt56sol}
& 18.94 & 20.63 & 22.05 \\
GPT-6 Astra~\oodcite{openai2026gpt6astra}
& 6.32 & 6.26 & 69.27 \\
\midrule[0.4pt]
\oodgroup{Monocular Depth VLMs} \\[1pt]
DepthLM-13B~\oodcite{yeh2026depthlm}
& -- & -- & 31.32 \\
DepthVLM-4B~\oodcite{yu2026unlocking}
& -- & -- & 40.08 \\
\midrule[0.4pt]
\oodgroup{Binocular VLMs} \\[1pt]
\rowcolor{stereooursbg}
\textbf{StereoVLM-2B}
& \stereogap{0.035\linewidth}4.95 & 5.95 & 68.92 \\
\rowcolor{stereooursbg}
\textbf{StereoVLM-8B}
& \stereogap{0.035\linewidth}\textbf{4.59} & \textbf{5.28} & \textbf{70.16} \\
\bottomrule[0.8pt]
\end{tabular}
\vspace{-3ex}
\end{wraptable}

Although StereoVLM is trained with explicit stereo supervision, it is
not restricted to binocular inputs. Prior work examines how self-supervised monocular depth estimators generalize across domains~\citep{yao2024monocular}. To evaluate the monocular capability retained by the model after stereo instruction tuning, we remove the right image and provide only the rectified left image. We also remove the stereo baseline from the question. The checkpoint, focal length, remaining question text, and decoding configuration are otherwise unchanged. This setting prevents
the model from using binocular correspondence and measures what can be
recovered from monocular appearance, object-scale priors, and pretrained
visual-language knowledge.

\Cref{tab:sup_single_view} shows that StereoVLM retains substantial
single-view depth and spatial reasoning capability. StereoVLM-2B and
StereoVLM-8B obtain point-level depth $\delta_1$ scores of 73.51 and
63.22, respectively, compared with 43.88 for the strongest
general-purpose baseline. StereoVLM-2B also reaches an object-depth
$\delta_1$ of 81.52. For oriented 3D grounding, StereoVLM-2B and
StereoVLM-8B achieve mIoU scores of 16.92 and 6.82, respectively.
On task-target grounding, their Joint@0.25 scores are 25.64 and 28.21,
respectively, exceeding the strongest general-purpose baseline at 7.69.

Compared with the stereo-input results in \cref{tab:stereo3d_results},
GPT-6 Astra's point-level depth $\delta_1$ drops from 75.96\% to 43.88\%,
while its object-depth $\delta_1$ falls from 93.40\% to 37.37\%
(\cref{tab:sup_single_view}). These correspond to decreases of 32.08 and
56.03 percentage points, respectively, indicating substantially higher
depth accuracy under the stereo evaluation setting. These results suggest that GPT-6 Astra benefits from stereo conditioning
for both point-level and object-level depth estimation, with a more
pronounced decline in object-depth accuracy under single-view input.

These results indicate that stereo-representation training does not
replace the monocular cues inherited from the pretrained model.
Instead, StereoVLM remains usable when only one image is available,
while the primary stereo evaluation measures the additional capability
obtained from a valid left--right pair. We omit correspondence and
disparity from this table because neither quantity is directly
observable from a single image.

\begin{table}[t]
\centering
\setlength{\belowcaptionskip}{5pt}
\setlength{\abovecaptionskip}{4pt}
\caption{Results on the main split of Stereo3D-Bench under single-view input. All models receive only the rectified left image, and the stereo baseline is removed from the question. For models with invalid depth
outputs, Depth $\delta_1$ is computed over all queries, treating invalid
outputs as failures.}
\label{tab:sup_single_view}
\fontsize{8}{9.5}\selectfont
\setlength{\tabcolsep}{0pt}
\renewcommand{\arraystretch}{0.98421}
\def\singleviewhead#1#2{%
  \shortstack[c]{{\fontsize{8}{9.5}\selectfont\strut #1}\\[-1pt]{\fontsize{7}{8}\selectfont\strut #2}}%
}
\def\singleviewnum#1{\makebox[2.25em][r]{#1}}
\begin{tabular}{@{}w{l}{0.305\linewidth}
  w{c}{0.17\linewidth}
  @{\hspace{0.015\linewidth}}
  *{3}{w{c}{0.17\linewidth}}@{}}
\toprule[0.8pt]
\multirow[c]{2}{*}[-\dimexpr(\aboverulesep+\belowrulesep+\cmidrulewidth)/2+5.7pt\relax]{Model}
& Geometry
& \multicolumn{3}{c}{Semantic} \\
\cmidrule(lr){2-2}\cmidrule(l){3-5}
& \singleviewhead{Depth}{$\delta_1$ $\uparrow$}
& \singleviewhead{Depth}{$\delta_1$ $\uparrow$}
& \singleviewhead{3D Ground.}{mIoU $\uparrow$}
& \singleviewhead{Target}{Joint@0.25 $\uparrow$} \\
\midrule
\multicolumn{5}{l}{\itshape General-Purpose VLMs} \\[2pt]
Qwen3-VL-2B~\citep{bai2025qwen3}
& \singleviewnum{19.80}
& \singleviewnum{28.56}
& \singleviewnum{0.06}
& \singleviewnum{0.00} \\
Qwen3-VL-8B~\citep{bai2025qwen3}
& \singleviewnum{12.77}
& \singleviewnum{13.97}
& \singleviewnum{0.20}
& \singleviewnum{0.00} \\
GPT-5.6 Sol~\citep{openai2026gpt56sol}
& \singleviewnum{17.25}
& \singleviewnum{29.62}
& \singleviewnum{0.18}
& \singleviewnum{0.00} \\
GPT-6 Astra~\citep{openai2026gpt6astra}
& \singleviewnum{43.88}
& \singleviewnum{37.37}
& \singleviewnum{7.02}
& \singleviewnum{7.69} \\
\midrule[0.3pt]
\multicolumn{5}{l}{\itshape Binocular VLMs} \\[2pt]
\textbf{StereoVLM-2B}
& \singleviewnum{\textbf{73.51}}
& \singleviewnum{\textbf{81.52}}
& \singleviewnum{\textbf{16.92}}
& \singleviewnum{25.64} \\
\textbf{StereoVLM-8B}
& \singleviewnum{63.22}
& \singleviewnum{53.33}
& \singleviewnum{6.82}
& \singleviewnum{\textbf{28.21}} \\
\bottomrule[0.8pt]
\end{tabular}
\end{table}

\subsection{Robustness to Extrinsic Calibration Errors}
\label{sec:sup_calibration}

Camera pose determines how 3D scene points project into an image~\citep{yan2024gsslam,yan2026pgocc}. We test how sensitive StereoVLM is to small errors in the relative pose of the stereo cameras. Starting from a rectified pair, we keep the left
image, query pixels, nominal focal length, nominal baseline, and
left-view depth references fixed, and perturb only the pose of the
right camera. The resulting right image is rendered under one of six
single-axis perturbations: rotations $R_x$, $R_y$, and $R_z$, or
translations $t_x$, $t_y$, and $t_z$. The experiment therefore
measures sensitivity to imperfect stereo alignment without changing
the scene or the queried left-view geometry.

The evaluation contains 400 base pairs, comprising 200 DROID pairs and
200 in-house robot pairs. Each pair contributes eight fixed left-image
query points. For every axis, we construct three perturbation levels,
giving 3,200 queried points per axis and level. The rotation magnitudes
are shared across the three rotational axes, and the translation
magnitudes are shared across the three translational axes. Averaged
over the evaluation set, L1, L2, and L3 correspond to rotations of
approximately $0.20^\circ$, $0.59^\circ$, and $0.98^\circ$, or
translations of 3.7\,mm, 11.1\,mm, and 18.4\,mm. These magnitudes are
selected per pair such that $R_x$ and $t_y$ induce mean vertical
residuals of approximately 1, 3, and 5 pixels. The other axes use the
same physical rotation or translation magnitude rather than being
independently adjusted to produce the same image-space displacement.

\begin{figure*}[t]
  \centering
  \includegraphics[width=\textwidth]
  {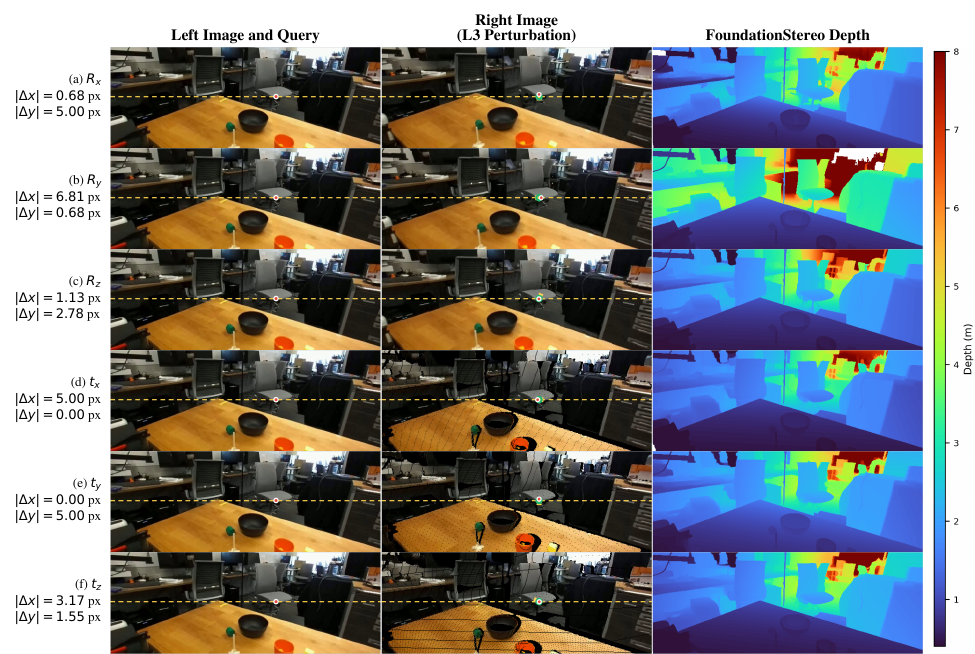}
  \vspace{-4ex}
  \caption{Qualitative effects of the strongest single-axis calibration
  perturbations. Each row shows the fixed left image and query, the
  right image after an L3 perturbation, and the metric depth produced
  by FoundationStereo from the perturbed pair. Green circles mark
  nominal correspondences, red points mark their locations after
  perturbation, and yellow arrows show the induced displacement. Row
  labels give the perturbation direction and its mean horizontal and
  vertical displacement. FoundationStereo receives the complete
  perturbed images; the common crop is applied only for display, and
  all depth maps share the same color scale. The displayed DROID
  example uses $0.99^\circ$ rotations and 21.2\,mm translations;
  perturbation magnitudes vary slightly with each pair's scene geometry.}
  \label{fig:sup_calibration_axes}
  \vspace{-2ex}
\end{figure*}

\Cref{fig:sup_calibration_axes} illustrates the strongest perturbation
for each axis. Each row shows the fixed left image, the perturbed right
image, and the metric depth produced by FoundationStereo from that
pair. The red point in the left image is the queried pixel. In the
right image, the green circle marks its nominal correspondence and the
red point marks its location after perturbation. The row labels report
the mean horizontal and vertical correspondence displacement induced
by the perturbation.

Rotations about $x$ and translations along $y$ primarily move
correspondences away from the original epipolar row. By contrast,
$R_y$ and $t_x$ primarily change horizontal position and therefore
alter apparent disparity. The effects of $R_z$ and $t_z$ depend on
image position and scene depth. FoundationStereo is run on the
complete perturbed pair before the common display crop is applied. Its
depth maps are included only as a qualitative diagnostic and do not
contribute to the StereoVLM results in
\cref{tab:sup_calibration}. Black regions in the translation examples are invalid areas caused by
depth-based forward warping.

We evaluate three outputs. Correspondence EPE is measured against the
right-image coordinate after perturbation. Direct depth is the metric
depth predicted by StereoVLM in response to the depth question.
Finally, we derive a depth estimate from the predicted right-image
coordinate using
\begin{equation}
  \hat d_i = x_i^l-\hat x_i^r,
  \qquad
  \hat Z_i^{\mathrm{corr}} =
  \frac{f_xB}{\hat d_i},
\end{equation}
where $f_x$ and $B$ are the nominal focal length and baseline.
Because a perturbed pair does not in general satisfy the nominal
rectified-stereo model, this quantity is a diagnostic of prediction
under the original calibration rather than a physically corrected
triangulation using the perturbed extrinsics. It is derived from the
predicted correspondence and is not a separately predicted disparity
output.

\Cref{tab:sup_calibration} gives the complete results for
StereoVLM-2B. Correspondence error increases as the perturbation
becomes stronger, from 2.48 pixels without perturbation to 4.93--7.13
pixels under the strongest perturbations. Direct metric depth is more
stable. Under rotational perturbations, depth $\delta_1$ remains
between 90.3\% and 91.1\%, close to the nominal value of 90.5\%.
Translation has a larger but still moderate effect: at L3, depth
$\delta_1$ ranges from 86.1\% to 87.8\%, and MAE remains below
20\,cm.

Depth derived under the nominal calibration follows the same broad
trend, decreasing from 90.0\% under the nominal condition to
88.0--89.2\% under the strongest rotations and 84.6--85.5\% under the
strongest translations. Meanwhile, correspondence EPE increases
consistently with perturbation magnitude. Thus, the model remains
numerically stable but does not fully adapt its correspondence
predictions to the perturbed geometry. Together with the single-view
results in \cref{sec:sup_single_view}, this suggests that direct depth combines binocular evidence with
monocular and nominal-stereo priors.

\begin{table*}[t]
\centering
\setlength{\abovecaptionskip}{4pt}
\setlength{\belowcaptionskip}{5pt}
\caption{Robustness of StereoVLM-2B to single-axis right-camera
perturbations. Values within each perturbed row are reported in
L1/L2/L3 order. Direct depth is predicted by StereoVLM.
Nominal-calibration depth is derived from the predicted correspondence
using $\hat Z=f_xB/(x_l-\hat x_r)$ without correcting for the perturbed
extrinsics.}
\label{tab:sup_calibration}
\fontsize{7.5}{9}\selectfont
\setlength{\tabcolsep}{0pt}
\renewcommand{\arraystretch}{1.03889}
\def\calnum#1{\makebox[7.4em][r]{#1}}
\begin{tabular}{@{}w{l}{0.18\linewidth}*{4}{w{c}{0.205\linewidth}}@{}}
\toprule[0.8pt]
\multirow[c]{2}{*}[-\dimexpr(\aboverulesep+\belowrulesep+\cmidrulewidth)/2\relax]{Condition}
& Correspondence
& \multicolumn{2}{c}{Direct Depth}
& Nominal Depth \\
\cmidrule(lr){2-2}\cmidrule(lr){3-4}\cmidrule(l){5-5}
& {\scriptsize EPE (px) $\downarrow$}
& {\scriptsize $\delta_1$ $\uparrow$}
& {\scriptsize MAE (cm) $\downarrow$}
& {\scriptsize $\delta_1$ $\uparrow$} \\
\midrule
\rowcolor{stereooursbg}
Nominal
& 2.48
& 90.5
& 16.5
& 90.0 \\
\midrule
$R_x$, L1/L2/L3
& \calnum{2.91 / 4.54 / 6.41}
& \calnum{90.6 / 90.7 / 90.6}
& \calnum{16.2 / 16.5 / 17.3}
& \calnum{90.2 / 89.8 / 89.2} \\
$R_y$, L1/L2/L3
& \calnum{2.93 / 4.97 / 7.13}
& \calnum{90.6 / 90.5 / 90.3}
& \calnum{16.7 / 17.8 / 17.1}
& \calnum{89.8 / 89.2 / 88.0} \\
$R_z$, L1/L2/L3
& \calnum{2.69 / 3.65 / 4.93}
& \calnum{90.6 / 91.1 / 91.0}
& \calnum{16.4 / 16.4 / 16.2}
& \calnum{89.2 / 89.4 / 89.1} \\
\midrule
$t_x$, L1/L2/L3
& \calnum{3.43 / 4.43 / 5.77}
& \calnum{87.9 / 87.8 / 87.1}
& \calnum{18.9 / 18.6 / 18.8}
& \calnum{87.1 / 85.8 / 84.6} \\
$t_y$, L1/L2/L3
& \calnum{3.67 / 5.14 / 7.00}
& \calnum{88.6 / 88.1 / 87.8}
& \calnum{18.7 / 18.1 / 19.3}
& \calnum{86.2 / 86.7 / 85.5} \\
$t_z$, L1/L2/L3
& \calnum{3.56 / 4.54 / 5.71}
& \calnum{87.8 / 87.9 / 86.1}
& \calnum{18.4 / 18.7 / 19.7}
& \calnum{86.6 / 85.8 / 85.4} \\
\bottomrule[0.8pt]
\end{tabular}
\end{table*}

\clearpage
\subsection{Sensitivity to Baseline Calibration}
\label{sec:sup_baseline_calibration}

We evaluate how StereoVLM-2B's directly generated metric depth answers
respond to the baseline specified in the prompt using 400 stereo pairs: 200 from DROID
and 200 from our in-house robot dataset. We keep the stereo images,
eight query pixels per pair, focal length, and model checkpoint fixed,
and replace the nominal baseline $B_0$ with $B=\alpha B_0$, where
$\alpha\in\{0.6,0.8,0.9,1.0,1.1,1.2,1.5\}$.

For each query point $i$, we measure the predicted depth ratio
\begin{equation}
  r_i(\alpha)=\frac{\hat Z_i(\alpha B_0)}{\hat Z_i(B_0)}.
\end{equation}
With image disparity fixed, the geometric relation $Z=f_xB/d$ predicts
ideal scaling $r_i(\alpha)=\alpha$. This serves as a reference for the
model's directly generated answers. We report the median ratio and its
10th--90th percentile range over all 1,600 query points in each domain,
without filtering samples.

\begin{figure}[htbp]
  \centering
  \setlength{\abovecaptionskip}{-2pt}
  \includegraphics[width=\textwidth]
  {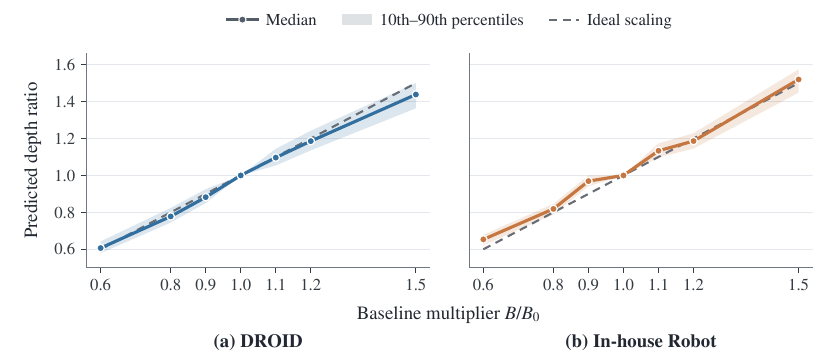}
  \caption{Sensitivity of StereoVLM-2B to the baseline supplied in the prompt
  on DROID and the in-house robot dataset. Solid curves show median
  predicted depth ratios; shaded bands indicate the 10th--90th percentile
  range. Dashed lines show ideal linear scaling. Images and focal length
  are fixed throughout.}
  \label{fig:sup_baseline_calibration}
\end{figure}

As shown in \Cref{fig:sup_baseline_calibration}, predicted depth increases
with the supplied baseline in both domains, with median ratios broadly
following the ideal scaling relation. These results suggest that the
model uses the supplied baseline for metric depth prediction.

\clearpage
\subsection{Additional Transfer Experiments}
\label{sec:sup_transfer}

\subsubsection{General Spatial Reasoning}
\label{sec:sup_spatial_transfer}

To separate the effects of spatial supervision and binocular input in
\cref{tab:stereo_data_ablation}, we compare three training-input conditions:
\emph{Correct Stereo} uses paired left and right images, \emph{Single Left}
uses only the left image, and \emph{Duplicated Left} copies the left image
into both input positions. All three share a reduced supervision set
covering depth estimation, object-level and task-target 3D grounding, and
other spatial tasks. Correspondence and disparity questions are excluded
from this shared set because Single Left lacks the required right-view
observation. \emph{StereoVLM} uses correctly paired images and additionally
includes correspondence and disparity supervision. We evaluate all variants
on the same three monocular and two video benchmarks with their standard
visual inputs.

\begin{table}[!htbp]
\centering
\setlength{\belowcaptionskip}{5pt}
\setlength{\abovecaptionskip}{4pt}
\caption{\raggedright Monocular and video spatial reasoning by training input and geometric supervision (Corr./Disp./Depth). Avg. is the unweighted mean of the five benchmarks; gains are relative to the w/o-stereo baseline.}
\label{tab:sup_spatial_transfer}
\fontsize{7}{8.5}\selectfont
\setlength{\tabcolsep}{0pt}
\renewcommand{\arraystretch}{1.15}
\def\transferhead#1{{\fontsize{6.5}{7.5}\selectfont #1}}
\def\transfernum#1{\makebox[2.25em][r]{#1}}
\def\transfergain#1#2{\transfernum{#1}\hspace{1pt}{\fontsize{6.5}{7.5}\selectfont\color{red!80!black}($\uparrow$\,#2)}}
\begin{tabular}{@{}
  w{l}{0.205\linewidth}
  *{3}{w{c}{0.048\linewidth}}
  @{\hspace{0.01\linewidth}}
  *{3}{w{c}{0.095\linewidth}}
  @{\hspace{0.01\linewidth}}
  *{2}{w{c}{0.095\linewidth}}
  w{c}{0.15\linewidth}
  @{}}
\toprule[0.8pt]
\multirow[c]{2}{*}[-\dimexpr(\aboverulesep+\belowrulesep+\cmidrulewidth)/2\relax]{\transferhead{Training input}}
& \multicolumn{3}{c}{\transferhead{Supervision}}
& \multicolumn{3}{c}{\transferhead{Monocular}}
& \multicolumn{2}{c}{\transferhead{Video}}
& \multirow[c]{2}{*}[-\dimexpr(\aboverulesep+\belowrulesep+\cmidrulewidth)/2\relax]{\transferhead{Avg. $\uparrow$}} \\
\cmidrule(lr){2-4}\cmidrule(lr){5-7}\cmidrule(lr){8-9}
& \transferhead{Corr.} & \transferhead{Disp.} & \transferhead{Depth}
& \transferhead{Omni3D} & \transferhead{DA-2K} & \transferhead{OmniSpatial}
& \transferhead{MindCube} & \transferhead{VSI-Bench} & \\
\midrule[0.4pt]
w/o stereo & -- & -- & --
& \transfernum{23.95} & \transfernum{73.79} & \transfernum{40.18}
& \transfernum{44.86} & \transfernum{56.94} & \transfernum{47.94} \\
Correct stereo\,$(\mathbf{I}_l,\mathbf{I}_r)$ & -- & -- & $\checkmark$
& \transfernum{30.09} & \transfernum{71.76} & \transfernum{44.10}
& \transfernum{47.81} & \transfernum{57.40} & \transfernum{50.23} \\
Single left\,$\mathbf{I}_l$ & -- & -- & $\checkmark$
& \transfernum{28.50} & \transfernum{80.66} & \transfernum{40.70}
& \transfernum{45.81} & \transfernum{55.39} & \transfernum{50.21} \\
Duplicated left\,$(\mathbf{I}_l,\mathbf{I}_l)$ & -- & -- & $\checkmark$
& \transfernum{32.78} & \transfernum{75.58} & \transfernum{41.81}
& \transfernum{47.05} & \transfernum{50.08} & \transfernum{49.46} \\
\midrule[0.4pt]
\noalign{\begingroup\color{stereooursbg}\hrule height9.35pt\endgroup\kern-9.35pt}
\textbf{StereoVLM-2B} & $\checkmark$ & $\checkmark$ & $\checkmark$
& \transfernum{32.16} & \transfernum{76.16} & \transfernum{42.66}
& \transfernum{48.57} & \transfernum{57.86} & \transfergain{\textbf{51.48}}{3.54} \\
\bottomrule[0.8pt]
\end{tabular}
\end{table}

\Cref{tab:sup_spatial_transfer} illustrates the roles of spatial supervision and training input.
With the shared supervision set, Correct Stereo, Single Left, and
Duplicated Left achieve averages of 50.23, 50.21, and 49.46, respectively,
all exceeding the 47.94 baseline without stereo instruction tuning.
Additional spatial supervision therefore benefits transfer even without
valid stereo pairs. Correct Stereo improves over Duplicated Left by
0.77 points, but performs similarly to Single Left on average, indicating
that paired images alone do not ensure stronger transfer.

With full geometric supervision, StereoVLM achieves the highest average
of 51.48, exceeding Correct Stereo by 1.25 points, and leads on both
MindCube and VSI-Bench. Although no configuration dominates every
benchmark, these results support combining correspondence, disparity,
and depth supervision to improve overall spatial-reasoning transfer.

\clearpage
\subsubsection{Embodied Control}
\label{sec:sup_embodied_control}

\begin{wraptable}{r}{0.51\textwidth}
\vspace{-3.5ex}
\centering
\setlength{\abovecaptionskip}{4pt}
\setlength{\belowcaptionskip}{5pt}
\caption{\raggedright Manipulation success (\%) by visual input and geometric supervision (Corr./Disp./Depth). Each configuration uses the same visual inputs during training and evaluation. The gain is relative to the best depth-supervised variant.}
\label{tab:sup_control_training_input}
\centering
\fontsize{7}{8.5}\selectfont
\setlength{\tabcolsep}{0pt}
\renewcommand{\arraystretch}{1.15}
\def\condhead#1{{\fontsize{6.5}{7.5}\selectfont #1}}
\def\condnum#1{\makebox[2.15em][r]{#1}}
\def\condgain#1#2{\condnum{#1}\hspace{1pt}{\fontsize{6.5}{7.5}\selectfont\color{red!80!black}($\uparrow$\,#2)}}
\begin{tabular}{@{}w{l}{0.40\linewidth}*{3}{w{c}{0.11\linewidth}}w{c}{0.27\linewidth}@{}}
\toprule[0.8pt]
\multirow[c]{2}{*}[-\dimexpr(\aboverulesep+\belowrulesep+\cmidrulewidth)/2\relax]{\condhead{Visual input}}
& \multicolumn{3}{c}{\condhead{Supervision}}
& \multirow[c]{2}{*}[-\dimexpr(\aboverulesep+\belowrulesep+\cmidrulewidth)/2\relax]{\condhead{Success $\uparrow$}} \\
\cmidrule(lr){2-4}
& \condhead{Corr.} & \condhead{Disp.} & \condhead{Depth} & \\
\midrule[0.4pt]
Correct stereo\,$(\mathbf{I}_l,\mathbf{I}_r)$ & -- & -- & $\checkmark$ & \condnum{38.2} \\
Single left\,$\mathbf{I}_l$ & -- & -- & $\checkmark$ & \condnum{37.7} \\
Duplicated left\,$(\mathbf{I}_l,\mathbf{I}_l)$ & -- & -- & $\checkmark$ & \condnum{36.3} \\
\midrule[0.4pt]
\noalign{\begingroup\color{stereooursbg}\hrule height9.35pt\endgroup\kern-9.35pt}
\textbf{StereoVLM-2B} & $\checkmark$ & $\checkmark$ & $\checkmark$ & \condgain{\textbf{42.4}}{4.2} \\
\bottomrule[0.8pt]
\end{tabular}
\vspace{-4ex}
\end{wraptable}

\textbf{Transfer of geometric supervision.}
We evaluate the configurations in \cref{sec:sup_spatial_transfer} on
downstream manipulation, using the same visual input configuration during
training and evaluation. When point-level geometric supervision is limited
to depth, the correct-stereo and single-left configurations achieve similar
success rates (38.2\% and 37.7\%; \cref{tab:sup_control_training_input}).
Within the correct-stereo configuration, adding correspondence and disparity
supervision raises success to 42.4\%, a gain of 4.2 percentage points.
This improvement supports explicitly supervising cross-view geometry
when learning representations for embodied control.

\begin{wraptable}{r}{0.51\textwidth}
\vspace{-3.5ex}
\centering
\setlength{\abovecaptionskip}{4pt}
\setlength{\belowcaptionskip}{5pt}
\caption{\raggedright Embodied-control success rates. Correct stereo and duplicated
left are inference conditions for the same stereo-policy checkpoint;
single left uses a separately trained policy. All policies follow the
same downstream training and evaluation protocol.}
\label{tab:sup_embodied_control}
\fontsize{7}{8.5}\selectfont
\setlength{\tabcolsep}{0pt}
\renewcommand{\arraystretch}{1.15}
\def\embodiednum#1{\makebox[2.25em][r]{#1}}
\def\embodiedhead#1{{\fontsize{6.5}{7.5}\selectfont #1}}
\begin{tabular}{@{}w{l}{0.43\linewidth}*{3}{w{c}{0.19\linewidth}}@{}}
\toprule[0.8pt]
\embodiedhead{VLM initialization}
& \embodiedhead{Stereo $\uparrow$}
& \embodiedhead{Dup. left $\uparrow$}
& \embodiedhead{Single left $\uparrow$} \\
\midrule[0.4pt]
Qwen3-VL-2B
& \embodiednum{34.9} & \embodiednum{31.6} & \embodiednum{36.4} \\
\ours-2B w/o stereo
& \embodiednum{38.1} & \embodiednum{\textbf{36.8}} & \embodiednum{39.5} \\
\midrule[0.4pt]
\rowcolor{stereooursbg}
\textbf{\ours-2B}
& \embodiednum{\textbf{42.4}} & \embodiednum{26.3} & \embodiednum{\textbf{40.0}} \\
\bottomrule[0.8pt]
\end{tabular}
\vspace{-3ex}
\end{wraptable}

\textbf{Dependence on binocular observations.} We further examine whether stereo representation learning transfers to
embodied control. For each VLM initialization, we train separate stereo
and single-left policies under the same downstream training protocol.
The stereo policy is evaluated with either the correct stereo pair or
the left image duplicated into the right-view slot; these two results
use the same policy checkpoint. The single-left policy is trained and
evaluated using only the left image. \Cref{tab:sup_embodied_control}
reports average task success rates.

Without stereo representation training, the separately trained
single-left policies outperform their stereo counterparts: 36.4\%
versus 34.9\% for Qwen3-VL, and 39.5\% versus 38.1\% for \ours-2B w/o Stereo Supervision.
Thus, adding a right view does not by itself improve control under
this training protocol.

With stereo representation training, the stereo policy reaches 42.4\%
success, the highest result in the table. This exceeds the separately
trained single-left policy initialized from the same VLM by 2.4
percentage points. Because the policies are trained separately, this
comparison measures the outcome of two training configurations, not
the effect of removing the right view at inference time.

For the stereo-trained policy, replacing the right image with a copy
of the left reduces success from 42.4\% to 26.3\%. This 16.1-point
drop is a within-checkpoint intervention: it preserves the two-image
input format while removing the valid stereo pairing. The result
shows that this policy is sensitive to the content of the right-view
slot, although the intervention alone cannot distinguish geometric
dependence from other effects of an out-of-distribution input.

\clearpage
\subsection{Qualitative Visual Results}
\label{sec:sup_qualitative}

\begin{figure*}[!ht]
  \centering
  \includegraphics[width=\textwidth]{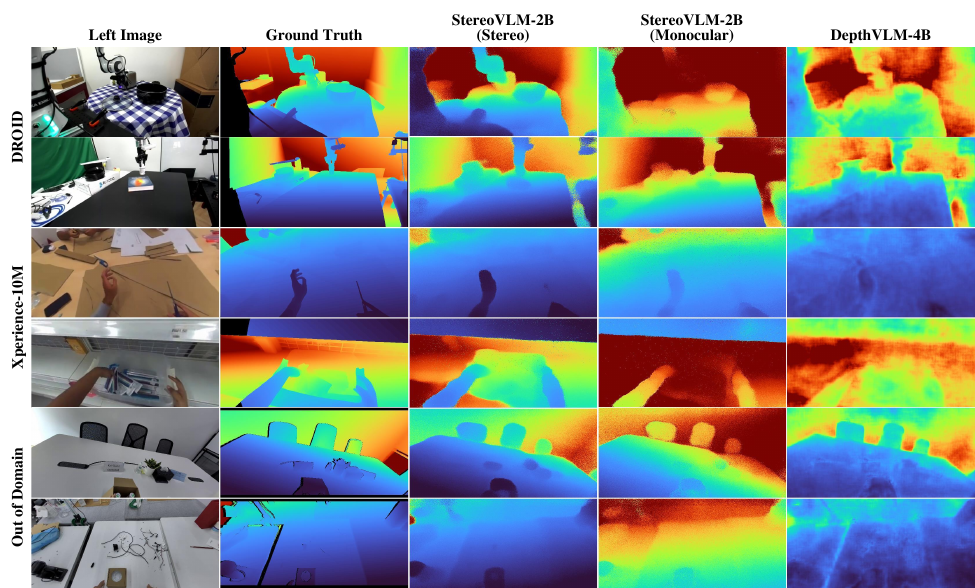}
  \vspace{-4ex}
  \caption{Qualitative dense depth results across three evaluation domains. Columns show the rectified left image, ground-truth depth, StereoVLM-2B with standard stereo input, the same StereoVLM-2B checkpoint with monocular input, and DepthVLM-4B. Rows are drawn from DROID, Xperience-10M, and the in-house OOD split.}
  \label{fig:sup_depth_results}
\end{figure*}

\Cref{fig:sup_depth_results} compares dense depth predictions across
DROID, Xperience-10M, and the in-house OOD split.
Alongside the standard stereo-input result, we show the same
StereoVLM-2B checkpoint with monocular input and the DepthVLM-4B
baseline for reference.

\begin{figure*}[!ht]
  \centering
  \includegraphics[width=\textwidth]
  {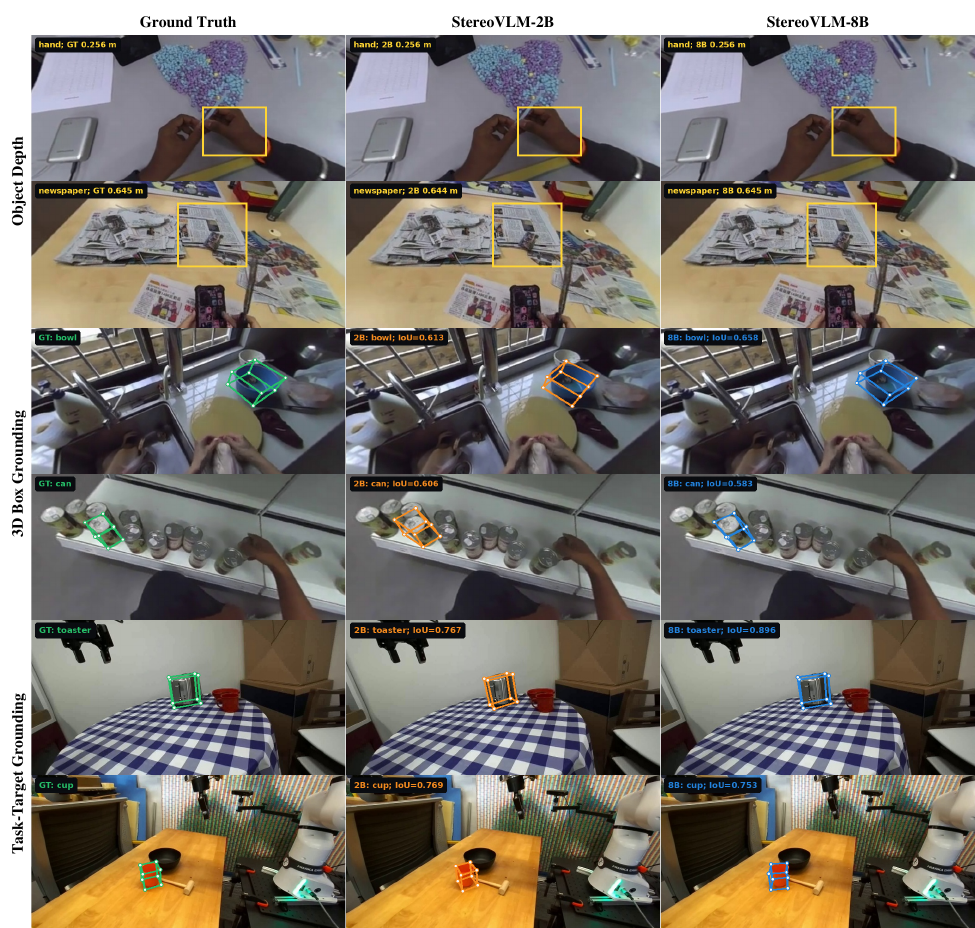}
  \vspace{-4ex}
  \caption{Qualitative semantic results under correct stereo input.
  Columns show the ground truth, StereoVLM-2B, and StereoVLM-8B.
  Object-depth examples report optical-axis depth in meters and use
  yellow boxes to indicate the queried objects. Semantic 3D OBB and
  task-target examples show oriented 3D boxes projected into the
  rectified-left image. Green, orange, and blue denote the ground truth,
  2B prediction, and 8B prediction, respectively. Depth and semantic
  OBB examples are from Xperience-10M, while task-target examples are
  from DROID. Only the left-view visualization is displayed; predictions
  are produced from the complete stereo pair.}
  \label{fig:sup_semantic_results}
\end{figure*}

\Cref{fig:sup_semantic_results} visualizes representative semantic
predictions under the standard stereo-input condition. Although only
the rectified left image is shown for readability, both models receive
the complete rectified left--right pair when producing each prediction.
The three row groups correspond to object metric depth, semantic
oriented 3D grounding, and task-target grounding.

For object depth, the yellow 2D box identifies the referred object and
the label reports its predicted optical-axis depth in meters. For
semantic 3D grounding, each oriented 3D box is projected into the
rectified-left image. Task-target grounding additionally requires the
model to identify the visible object specified by the manipulation
instruction before predicting its 3D box. Green marks the reference; orange and blue mark StereoVLM-2B and
StereoVLM-8B predictions, respectively.

In these examples, both model scales recover object depth closely and
produce 3D boxes with similar centers, extents, and orientations to the
references. The task-target examples further illustrate that the models
can connect a manipulation instruction to the appropriate scene object
and localize it in 3D. These examples are qualitative illustrations;
aggregate performance and controlled input comparisons are reported in
the main experiments.

\clearpage

\clearpage
\subsection{Comparison with Modular Stereo--VLM Pipelines}
\label{sec:sup_modular}

We evaluate whether a general-purpose VLM equipped with an external
stereo estimator can achieve comparable object-level spatial reasoning
to \ours. We construct two modular baselines using
FoundationStereo~\citep{wen2025foundationstereo} and a frozen
Qwen3-VL-2B-Instruct~\citep{bai2025qwen3}, and compare their semantic
accuracy and end-to-end inference latency with \ours-2B.

\paragraph{Baseline construction.}
The two baselines differ in how stereo geometry is provided to the VLM.
The \emph{localization-based} pipeline first queries the VLM to localize
the referred object in the left image. It then aggregates
FoundationStereo-predicted disparities within the localized region using
a fixed central-window median and disparity-based filtering. The selected
pixels are back-projected into 3D, and a PCA-based heuristic estimates an
approximate oriented bounding box of the visible surface. These geometric
measurements are then provided to the VLM to generate the final answer.
The \emph{direct-input} pipeline instead provides the stereo RGB
pair, a predicted depth visualization, an $8\times20$ disparity/depth
grid, and camera calibration to the VLM in a single call. Neither
baseline receives ground-truth answers, bounding boxes, or masks
at inference time. \ours-2B directly answers each question from the
stereo RGB pair and the task prompt, including the applicable camera
calibration, in one VLM call.

\begin{table}[htbp]
\centering
\setlength{\abovecaptionskip}{4pt}
\setlength{\belowcaptionskip}{5pt}
\caption{Accuracy and time efficiency of modular stereo--VLM pipelines
and \ours-2B.
FS denotes FoundationStereo; both modular pipelines use frozen
Qwen3-VL-2B. Semantic accuracy (\%) is evaluated on Stereo3D-Bench
using the metrics in \cref{tab:stereo3d_results}; Avg. is the unweighted
mean of the five task scores. Time efficiency is measured by mean
latency on an NVIDIA H200 at batch size 1, including
FoundationStereo inference for both modular pipelines.}
\label{tab:sup_modular_accuracy}
\fontsize{7.5}{9}\selectfont
\setlength{\tabcolsep}{0pt}
\renewcommand{\arraystretch}{1.03889}
\def\modularnum#1{\makebox[2.25em][r]{#1}}
\def\modularlatency#1{\makebox[2.75em][r]{#1}}
\begin{tabular}{@{}w{l}{0.24\linewidth}*{5}{w{c}{0.105\linewidth}}
  w{c}{0.105\linewidth}w{c}{0.13\linewidth}@{}}
\toprule[0.8pt]
\multirow[c]{2}{*}[-\dimexpr(\aboverulesep+\belowrulesep+\cmidrulewidth)/2\relax]{Model}
& \multicolumn{6}{c}{Semantic Accuracy (\%)}
& Latency \\
\cmidrule(lr){2-7}\cmidrule(l){8-8}
& Avg. $\uparrow$ & Corr. $\uparrow$ & Disp. $\uparrow$
& Depth $\uparrow$ & 3D Box $\uparrow$ & Target $\uparrow$
& Mean (s) $\downarrow$ \\
\midrule
\multicolumn{8}{@{}l}{\itshape Modular pipelines (FS + frozen Qwen3-VL-2B)} \\
Localization-based
& \modularnum{35.81} & \modularnum{43.75} & \modularnum{39.70}
& \modularnum{77.67} & \modularnum{12.80} & \modularnum{5.13}
& \modularlatency{2.212} \\
Direct-input
& \modularnum{5.07} & \modularnum{7.51} & \modularnum{0.52}
& \modularnum{17.03} & \modularnum{0.31} & \modularnum{0.00}
& \modularlatency{0.920} \\
\midrule
\multicolumn{8}{@{}l}{\itshape Binocular VLM} \\
\rowcolor{stereooursbg}
\textbf{StereoVLM-2B}
& \modularnum{\textbf{64.43}} & \modularnum{\textbf{86.06}}
& \modularnum{\textbf{53.92}} & \modularnum{\textbf{97.39}}
& \modularnum{\textbf{36.06}} & \modularnum{\textbf{48.72}}
& \modularlatency{\textbf{0.643}} \\
\bottomrule[0.8pt]
\end{tabular}
\end{table}

\paragraph{Accuracy and efficiency.}
Compared with modular pipelines that explicitly estimate geometry and
pass it to a frozen VLM, \ours predicts language-referred object geometry
and task targets directly from stereo images in a single VLM call.
This removes the separate stereo estimator and intermediate region-pooling
or geometry-encoding stages. As shown in \cref{tab:sup_modular_accuracy},
\ours-2B achieves higher accuracy on all five semantic tasks and lower
end-to-end latency in this comparison, with $3.44\times$ and $1.43\times$
speedups over the localization-based and direct-input pipelines,
respectively.

%% file: iclr2027_conference.bib
@inproceedings{brazil2023omni3d,
  title={{Omni3D}: A Large Benchmark and Model for {3D} Object Detection in the Wild},
  author={Brazil, Garrick and Kumar, Abhinav and Straub, Julian and Ravi, Nikhila and Johnson, Justin and Gkioxari, Georgia},
  booktitle={CVPR},
  year={2023}
}

@inproceedings{yang2024depthanythingv2,
  title={Depth Anything {V2}},
  author={Yang, Lihe and Kang, Bingyi and Huang, Zilong and Xu, Xiaogang and Feng, Jiashi and Zhao, Hengshuang},
  booktitle={NeurIPS},
  year={2024}
}

@inproceedings{chen2024spatialvlm,
  title={Spatialvlm: Endowing vision-language models with spatial reasoning capabilities},
  author={Chen, Boyuan and Xu, Zhuo and Kirmani, Sean and Ichter, Brain and Sadigh, Dorsa and Guibas, Leonidas and Xia, Fei},
  booktitle={CVPR},
  year={2024}
}

@article{deng2025stereovla,
  title={Stereovla: Enhancing vision-language-action models with stereo vision},
  author={Deng, Shengliang and Yan, Mi and Zheng, Yixin and Su, Jiayi and Zhang, Wenhao and Zeng, Yitao and Zhao, Xiaoguang and Cui, Heming and Zhang, Zhizheng and Wang, He},
  journal={arXiv preprint arXiv:2512.21970},
  year={2025}
}

@inproceedings{wen2025foundationstereo,
  title={Foundationstereo: Zero-shot stereo matching},
  author={Wen, Bowen and Trepte, Matthew and Aribido, Joseph and Kautz, Jan and Gallo, Orazio and Birchfield, Stan},
  booktitle={CVPR},
  year={2025},
  organization={IEEE}
}

@inproceedings{yang2025thinking,
  title={Thinking in space: How multimodal large language models see, remember, and recall spaces},
  author={Yang, Jihan and Yang, Shusheng and Gupta, Anjali W and Han, Rilyn and Fei-Fei, Li and Xie, Saining},
  booktitle={CVPR},
  year={2025},
  organization={IEEE}
}

@article{cheng2024spatialrgpt,
  title={Spatialrgpt: Grounded spatial reasoning in vision-language models},
  author={Cheng, An-Chieh and Yin, Hongxu and Fu, Yang and Guo, Qiushan and Yang, Ruihan and Kautz, Jan and Wang, Xiaolong and Liu, Sifei},
  journal={NeurlIPS},
  year={2024}
}

@article{hong20233dllm,
  title={3d-llm: Injecting the 3d world into large language models},
  author={Hong, Yining and Zhen, Haoyu and Chen, Peihao and Zheng, Shuhong and Du, Yilun and Chen, Zhenfang and Gan, Chuang},
  journal={NeurlIPS},
  year={2023}
}

@inproceedings{lipson2021raft,
  title={Raft-stereo: Multilevel recurrent field transforms for stereo matching},
  author={Lipson, Lahav and Teed, Zachary and Deng, Jia},
  booktitle={3DV},
  year={2021},
  organization={IEEE}
}

@inproceedings{xu2023iterative,
  title={Iterative geometry encoding volume for stereo matching},
  author={Xu, Gangwei and Wang, Xianqi and Ding, Xiaohuan and Yang, Xin},
  booktitle={CVPR},
  year={2023},
  organization={IEEE}
}

@article{alayrac2022flamingo,
  title={Flamingo: a visual language model for few-shot learning},
  author={Alayrac, Jean-Baptiste and Donahue, Jeff and Luc, Pauline and Miech, Antoine and Barr, Iain and Hasson, Yana and Lenc, Karel and Mensch, Arthur and Millican, Katherine and Reynolds, Malcolm and others},
  journal={NeurlIPS},
  year={2022}
}

@article{driess2023palm,
  title={Palm-e: An embodied multimodal language model},
  author={Driess, Danny and Xia, Fei and Sajjadi, Mehdi SM and Lynch, Corey and Chowdhery, Aakanksha and Ichter, Brian and Wahid, Ayzaan and Tompson, Jonathan and Vuong, Quan and Yu, Tianhe and others},
  journal={arXiv preprint arXiv:2303.03378},
  year={2023}
}

@article{zhang2026revsi,
  title={ReVSI: Rebuilding Visual Spatial Intelligence Evaluation for Accurate Assessment of VLM 3D Reasoning},
  author={Zhang, Yiming and Chen, Jiacheng and Tan, Jiaqi and Mao, Yongsen and Chen, Wenhu and Chang, Angel X.},
  journal={arXiv preprint arXiv:2604.24300},
  year={2026}
}

@inproceedings{yeh2026depthlm,
  title={Depthlm: Metric depth from vision language models},
  author={Yeh, Ching-Feng and Xu, Hu and Liu, Zhuang and Meyer, Gregory P and Lei, Xinjie and Zhao, Changsheng and Li, Shang-Wen and Chandra, Vikas and Shi, Yangyang and others},
  booktitle={ICLR},
  year={2026}
}

@article{yu2026unlocking,
  title={Unlocking Dense Metric Depth Estimation in VLMs},
  author={Yu, Hanxun and Qu, Xuan and Wang, Yuxin and Zhu, Jianke and Ke, Lei},
  journal={arXiv preprint arXiv:2605.15876},
  year={2026}
}

@article{nasiriany2024robocasa,
  title={{RoboCasa}: Large-scale simulation of everyday tasks for generalist robots},
  author={Nasiriany, Soroush and Maddukuri, Abhiram and Zhang, Lance and Parikh, Adeet and Lo, Aaron and Joshi, Abhishek and Mandlekar, Ajay and Zhu, Yuke},
  journal={arXiv preprint arXiv:2406.02523},
  year={2024}
}

@article{bai2025qwen3,
  title={Qwen3-vl technical report},
  author={Bai, Shuai and Cai, Yuxuan and Chen, Ruizhe and Chen, Keqin and Chen, Xionghui and Cheng, Zesen and Deng, Lianghao and Ding, Wei and Gao, Chang and Ge, Chunjiang and others},
  journal={arXiv preprint arXiv:2511.21631},
  year={2025}
}

@article{khazatsky2024droid,
  title={{DROID}: A large-scale in-the-wild robot manipulation dataset},
  author={Khazatsky, Alexander and Pertsch, Karl and Nair, Suraj and Balakrishna, Ashwin and Dasari, Sudeep and Karamcheti, Siddharth and Nasiriany, Soroush and Srirama, Mohan Kumar and Chen, Lawrence Yunliang and Ellis, Kirsty and others},
  journal={arXiv preprint arXiv:2403.12945},
  year={2024}
}

@misc{xperience_10m,
  title={{Xperience-10M}: A Large-Scale Egocentric Multimodal Dataset with Structured 3D/4D Annotations},
  author={Ropedia},
  year={2026},
  publisher={Hugging Face},
  note={Dataset}
}

@misc{openai2026gpt56sol,
  author       = {{OpenAI}},
  title        = {{GPT-5.6 Sol}},
  year         = {2026},
  howpublished = {OpenAI API model documentation},
  url          = {https://developers.openai.com/api/docs/models/gpt-5.6-sol},
  note         = {Accessed: 2026-09-24}
}

@misc{openai2026gpt6astra,
  author       = {{OpenAI}},
  title        = {{GPT-6 Astra}},
  year         = {2026},
  howpublished = {OpenAI API model documentation},
  url          = {https://developers.openai.com/api/docs/models/gpt-6-astra},
  note         = {Accessed: 2026-09-24}
}

@inproceedings{carion2026sam,
  title={Sam 3: Segment anything with concepts},
  author={Carion, Nicolas and Gustafson, Laura and Hu, Yuan-Ting and Debnath, Shoubhik and Hu, Ronghang and Suris Coll-Vinent, Didac and Ryali, Chaitanya and Alwala, Kalyan Vasudev and Khedr, Haitham and Huang, Andrew and others},
  booktitle={ICLR},
  year={2026}
}

@inproceedings{lian2025describe,
  title={Describe anything: Detailed localized image and video captioning},
  author={Lian, Long and Ding, Yifan and Ge, Yunhao and Liu, Sifei and Mao, Hanzi and Li, Boyi and Pavone, Marco and Liu, Ming-Yu and Darrell, Trevor and Yala, Adam and others},
  booktitle={ICCV},
  year={2025}
}

@article{wang2025n3d,
  title={N3d-vlm: Native 3d grounding enables accurate spatial reasoning in vision-language models},
  author={Wang, Yuxin and Ke, Lei and Zhang, Boqiang and Qu, Tianyuan and Yu, Hanxun and Huang, Zhenpeng and Yu, Meng and Xu, Dan and Yu, Dong},
  journal={arXiv preprint arXiv:2512.16561},
  year={2025}
}

@inproceedings{min2025s,
  title={S 2 m 2: scalable stereo matching model for reliable depth estimation},
  author={Min, Junhong and Jeon, Youngpil and Kim, Jimin and Choi, Minyong},
  booktitle={ICCV},
  year={2025},
  organization={IEEE}
}

@article{kim2024openvla,
  title={Openvla: An open-source vision-language-action model},
  author={Kim, Moo Jin and Pertsch, Karl and Karamcheti, Siddharth and Xiao, Ted and Balakrishna, Ashwin and Nair, Suraj and Rafailov, Rafael and Foster, Ethan and Lam, Grace and Sanketi, Pannag and others},
  journal={arXiv preprint arXiv:2406.09246},
  year={2024}
}

@article{brohan2023rt,
  title={Rt-2: Vision-language-action models transfer web knowledge to robotic control},
  author={Brohan, Anthony and Brown, Noah and Carbajal, Justice and Chebotar, Yevgen and Chen, Xi and Choromanski, Krzysztof and Ding, Tianli and Driess, Danny and Dubey, Avinava and Finn, Chelsea and others},
  journal={arXiv preprint arXiv:2307.15818},
  year={2023}
}

@article{han2026stereopolicy,
  title={StereoPolicy: Improving Robotic Manipulation Policies via Stereo Perception},
  author={Han, Evans and Jiang, Yunfan and Wang, Yingke and Xiao, Haoyue and Huang, Huang and Xie, Jianwen and Wu, Jiajun and Fei-Fei, Li and Zhang, Ruohan},
  journal={arXiv preprint arXiv:2605.09989},
  year={2026}
}

@article{wu2026eatr,
  title={EATR-Stereo: Embodiment-Aware Routing of Paired Stereo Evidence for Humanoid Vision-Language-Action Control},
  author={Wu, Songwei and Zhao, Rui and Yang, Fan and Nie, Zhongqiang and Jiang, Zhiduo and Sun, Wandong and Li, Yuwei and Liu, Yang and Liu, Hong},
  journal={arXiv preprint arXiv:2608.17453},
  year={2026}
}

@inproceedings{zhou2026vlbiman,
  title={Vlbiman: Vision-language anchored one-shot demonstration enables generalizable bimanual robotic manipulation},
  author={Zhou, Huayi and Jia, Kui},
  booktitle={ICLR},
  year={2026}
}

@article{cai2026vlm3,
  title={VLM3: Vision Language Models Are Native 3D Learners},
  author={Cai, Zhipeng and Liu, Zhuang and Xiong, Yunyang and Liu, Zechun and Chandra, Vikas and Shi, Yangyang},
  journal={arXiv preprint arXiv:2605.30561},
  year={2026}
}

@article{bjorck2025gr00t,
  title={Gr00t n1: An open foundation model for generalist humanoid robots},
  author={Bjorck, Johan and Casta{\~n}eda, Fernando and Cherniadev, Nikita and Da, Xingye and Ding, Runyu and Fan, Linxi and Fang, Yu and Fox, Dieter and Hu, Fengyuan and Huang, Spencer and others},
  journal={arXiv preprint arXiv:2503.14734},
  year={2025}
}

@inproceedings{jia2026omnispatial,
  title={Omnispatial: Towards comprehensive spatial reasoning benchmark for vision language models},
  author={Jia, Mengdi and Qi, Zekun and Zhang, Shaochen and Zhang, Wenyao and Yu, Xinqiang and He, Jiawei and Wang, He and Yi, Li},
  booktitle={ICLR},
  year={2026}
}

@inproceedings{wang2025spatial,
  title={Spatial mental modeling from limited views},
  author={Wang, Qineng and Yin, Baiqiao and Zhang, Pingyue and Zhang, Jianshu and Wang, Kangrui and Wang, Zihan and Zhang, Jieyu and Chandrasegaran, Keshigeyan and Liu, Han and Krishna, Ranjay and others},
  booktitle={ICLR},
  year={2025}
}

@article{zhou2026si0,
  title={{Si0}: The capabilities that transfer, perceived through the robot's eyes},
  author={Zhou, Yanning and Yuan, Wentao and Yan, Chi and Wang, Honghui and Zhao, Guoyang and Qiu, Feng and Wang, Kewei and Ge, Yixiao},
  journal={XPENG Robotics Blog},
  year={2026},
  url={https://xpeng-robotics.github.io/si0/}
}

@inproceedings{yao2024monocular,
  title={Improving Domain Generalization in Self-Supervised Monocular Depth Estimation via Stabilized Adversarial Training},
  author={Yao, Yuanqi and Wu, Gang and Jiang, Kui and Liu, Siao and Kuai, Jian and Liu, Xianming and Jiang, Junjun},
  booktitle={ECCV},
  year={2024}
}

@inproceedings{zhao2026geoprop,
  title={{GeoProp}: Grounding Robot State in Vision for Generalist Manipulation},
  author={Zhao, Guoyang and Qian, Quanhao and Zhang, Gongjie and Li, Wenhao and Wang, Jiuniu and Lu, Xiaowei and Zhao, Deli and Xu, Ran},
  booktitle={CoRL},
  year={2026}
}

@inproceedings{yan2024gsslam,
  title={{GS-SLAM}: Dense Visual {SLAM} with {3D} Gaussian Splatting},
  author={Yan, Chi and Qu, Delin and Xu, Dan and Zhao, Bin and Wang, Zhigang and Wang, Dong and Li, Xuelong},
  booktitle={CVPR},
  year={2024}
}

@inproceedings{yan2026pgocc,
  title={Progressive Gaussian Transformer with Anisotropy-aware Sampling for Open Vocabulary Occupancy Prediction},
  author={Yan, Chi and Xu, Dan},
  booktitle={ICLR},
  year={2026}
}

@article{cao2026spatialspeak,
  title={{SpatialSpeak}: {QA}-Native Reconstruction with Local and Global Context for Spatial Chain-of-Thought Reasoning},
  author={Cao, Yang and Zhang, Jiaxin and Chen, Dave Zhenyu and Zhong, Yingji and Gao, Ruiyuan and Hong, Lanqing and Xu, Dan},
  journal={arXiv preprint arXiv:2609.33616},
  year={2026}
}
